%% file: main.tex
\pdfoutput=1

\documentclass[11pt]{article}

\usepackage{ACL2023}

\usepackage{times}
\usepackage{latexsym}

\usepackage[T1]{fontenc}

\usepackage[utf8]{inputenc}

\usepackage{microtype}
\usepackage{textcomp}

\usepackage{inconsolata}

\usepackage{amsmath}
\usepackage{amssymb}
\usepackage{graphicx}

\usepackage{times}
\usepackage{latexsym}
\usepackage{amsmath}
\usepackage{amssymb}
\usepackage{enumitem}

\usepackage{enumitem}
\usepackage{graphicx} 
\usepackage{graphicx}

\usepackage{graphicx}
\usepackage{booktabs}
\usepackage{amsmath}
\usepackage{multirow}
\usepackage{colortbl}
\usepackage{array}
\usepackage{xcolor}
\usepackage{pgfplots}
\usepackage{tikz}
\usepackage{amsmath}
\usepackage{graphicx}
\usetikzlibrary{positioning, backgrounds}
\usepackage{hyperref}

\usepackage{pgfplots}
\usepackage{pgfplotstable}
\usepackage{graphicx}
\usepackage{caption}
\usepackage{pgfkeys}
\pgfplotsset{compat=1.17}
\usepackage{adjustbox}

\pgfplotsset{compat=1.18}
\definecolor{baselinegold}{HTML}{DAA520}  

\usepackage{float}
\usepackage{subcaption}  

\newenvironment{itemize*}
    {\begin{itemize}%
      \setlength{\itemsep}{5pt}%
      \setlength{\parskip}{0pt}}%
    {\end{itemize}}

\renewcommand{\ddag}{\raisebox{0.2ex}{$^{\ddagger}$}} 

\definecolor{myproto}{HTML}{f3752a} 

\newcommand{\ssec}[1]{\S~\ref{#1}}

\definecolor{categorycolor}{HTML}{FBD5A7}   
\definecolor{rfcolor}{HTML}{FFA994}         
\definecolor{stepcolor}{HTML}{8ED2C9}       

\newcommand{\highlightcat}[1]{%
  \begingroup
  \setlength{\fboxsep}{0pt}%
  \colorbox{categorycolor}{\strut\textbf{#1}}%
  \endgroup
}

\newcommand{\highlightrf}[1]{%
  \begingroup
  \setlength{\fboxsep}{0pt}%
  \colorbox{rfcolor}{\strut\textbf{#1}}%
  \endgroup
}

\newcommand{\highlightstep}[1]{%
  \begingroup
  \setlength{\fboxsep}{0pt}%
  \colorbox{stepcolor}{\strut\textbf{#1}}%
  \endgroup
}

\newcommand{\rrl}[1]{\textsc{\MakeLowercase{#1}}}

\definecolor{segmentgray}{HTML}{e6e6e6}       
\definecolor{catcolor}{HTML}{FBD5A7}          
\definecolor{rfcolor}{HTML}{FFA994}           
\definecolor{attrcolor}{HTML}{8ED2C9}         

\newcommand{\highlightc}[1]{\begingroup\setlength{\fboxsep}{1pt}\colorbox{catcolor}{#1}\endgroup}
\newcommand{\highlightr}[1]{\begingroup\setlength{\fboxsep}{1pt}\colorbox{rfcolor}{#1}\endgroup}
\newcommand{\highlightat}[1]{\begingroup\setlength{\fboxsep}{1pt}\colorbox{attrcolor}{#1}\endgroup}

\usepackage[normalem]{ulem} 

\definecolor{row1}{HTML}{f9f1d9} 
\definecolor{row2}{HTML}{d5e9e1} 
\definecolor{row3}{HTML}{faeaea} 

\newcommand{\sdag}{\textsuperscript{\tiny \dag}}
\newcommand{\sddag}{\textsuperscript{\tiny \ddag}}

\usepackage{tcolorbox}

\usepackage{tabularx,siunitx,xcolor,booktabs,ragged2e}
\newcolumntype{Y}{>{\RaggedRight\arraybackslash}X} 
\newcommand{\pos}[1]{\cellcolor{green!6}\textcolor{green!50!black}{\(\uparrow\)~#1}}
\newcommand{\negat}[1]{\cellcolor{red!6}\textcolor{red!70!black}{\(\downarrow\)~#1}}

\usepackage{graphicx,xcolor}

\title{Coupling Local Context and Global Semantic Prototypes via a Hierarchical Architecture for Rhetorical Roles Labeling}

\author{
  \textbf{Anas Belfathi}\textsuperscript{1}\quad
  \textbf{Nicolas Hernandez}\textsuperscript{1}\quad
  \textbf{Laura Monceaux}\textsuperscript{1}\quad
  \textbf{Warren Bonnard}\textsuperscript{2} \\
  \textbf{Mary Catherine Lavissiere}\textsuperscript{1}\quad
  \textbf{Christine Jacquin}\textsuperscript{1}\quad
  \textbf{Richard Dufour}\textsuperscript{1} \\
  \textsuperscript{1} Nantes Université, École Centrale Nantes, CNRS, LS2N, UMR 6004, F-44000 Nantes, France \\
  \textsuperscript{2} University of Lorraine, France \\
  \textbf{Correspondence:} \href{mailto:anas.belfathi@univ-nantes.fr}{anas.belfathi@univ-nantes.fr}
}

\begin{document}
\maketitle
\begin{abstract}

Rhetorical Role Labeling (RRL) identifies the functional role of each sentence in a document, a key task for discourse understanding in domains such as law and medicine. While hierarchical models capture local dependencies effectively, they are limited in modeling global, corpus-level features. To address this limitation, we propose two prototype-based methods that integrate local context with global representations. \textbf{Prototype-Based Regularization (PBR)} learns soft prototypes through a distance-based auxiliary loss to structure the latent space, while \textbf{Prototype-Conditioned Modulation (PCM)} constructs corpus-level prototypes and injects them during training and inference.
Given the scarcity of RRL resources, we introduce \textsc{SCOTUS-Law}, the first dataset of U.S. Supreme Court opinions annotated with rhetorical roles at three levels of granularity: \textit{category}, \textit{rhetorical function}, and \textit{step}. Experiments on legal, medical, and scientific benchmarks show consistent improvements over strong baselines, with $\sim4$ Macro-F1 gains on low-frequency roles. We further analyze the implications in the era of Large Language Models and complement our findings with expert evaluation.

\end{abstract}

\section{Introduction}

\input{figures/example}

Rhetorical Role Labeling (RRL) is the task of classifying each sentence according to its semantic role within a document. Since a sentence’s meaning is often shaped by its surrounding context, RRL is particularly useful in structured texts such as legal cases. Identifying rhetorical components (e.g., \rrl{Announcing} or \rrl{Analysis}; see Figure~\ref{fig:exemple}) is useful for downstream tasks such as information retrieval~\cite{neves-etal-2019-evaluation, safder2019bibliometric} and document summarization~\cite{kalamkar-etal-2022-corpus, muhammed2024impact}.

Initially, RRL was framed as a sentence-level classification problem, overlooking contextual dependencies between sentences~\cite{walker2019automatic}. Later,~\citet{brack2022cross} modeled it as sequence labeling with hierarchical architectures to capture intra-document dependencies and represent local context more effectively. This approach has since become the de facto standard in recent RRL studies~\cite{kalamkar-etal-2022-corpus, bhattacharya_deeprhole_2023, nigam-etal-2025-legalseg}.

The challenge is that these architectures fail to capture global features shared across documents. We argue that leveraging this information could help resolve ambiguities between closely related roles, as noted by~\citet{kalamkar-etal-2022-corpus}. In this context, prototype learning~\cite{snell2017} serves as a principled way to address this limitation by learning global representations that serve as semantic anchors for each label. This approach has shown strong performance across various NLP tasks, including named entity recognition~\cite{huang-etal-2023-pram}, relation classification~\cite{yu-etal-2022-dependency}, and legal citation prediction~\cite{luo-etal-2023-prototype}.

Building on these insights, we study how local context and global representations can be combined through semantic prototypes. To our knowledge, no prior work has pursued this objective in the context of RRL, particularly within a hierarchical framework.
Our main contributions are: 

\vspace{-0.8em}

\begin{itemize*}
    \item We introduce two semantic prototype-based methods: (i)
    \textbf{Prototype-Based Regularization (PBR)}, which aligns sentence embeddings with prototypes via an auxiliary distance-based loss; 
    and (ii) \textbf{Prototype-Conditioned Modulation (PCM)}, which builds a priori prototypes from the corpus and injects them through dedicated modules during both training and inference.

    \item We release \textsc{SCOTUS-Law}, the first manually annotated corpus of U.S. Supreme Court opinions segmented into rhetorical roles at three levels of granularity (see Figure~\ref{fig:exemple}).

    \item We assess generalizability of our approach on seven benchmarks: legal datasets with long documents, and medical/scientific datasets with short abstracts.

    \item To position our work within the recent era of Large Language Models (LLMs), we evaluate three open-source models and complement this with a manual expert assessment to explain model behaviors.
\end{itemize*}

\vspace{-0.8em}

We release both our code and dataset under an open-source license\footnote{\url{https://github.com/AnasBelfathi/RRL-HierProto}}.

\section{Related Works}

\subsection{Rhetorical Role Labeling Approaches}

The story of RRL began with traditional machine learning using hand-crafted features~\cite{ruch2007using, mcknight2003categorization, lin-etal-2006-generative}. Neural architectures marked a major shift, with BERT-based models capturing contextual dependencies~\cite{cohan-etal-2019-pretrained, devlin-etal-2019-bert}. Recent methods build on this by adopting hierarchical architectures~\cite{jin-szolovits-2018-hierarchical, brack2024sequential}, encoding documents at multiple levels to produce contextualized sentence representations.
More broadly, RRL is related to research that labels the rhetorical/argumentative function of sentences in discourse, such as Argumentative Zoning~\cite{teufel-etal-2009-towards}, as well as discourse-level frameworks like RST that model rhetorical relations between text spans~\cite{feng-hirst-2012-text}.
More recent work enriches these representations through contrastive learning~\cite{t-y-s-s-etal-2024-mind}, curriculum learning~\cite{t-y-s-s-etal-2024-hiculr}, and improved pretraining objectives~\cite{belfathi-selective-2025}, moving beyond hierarchical encoding toward deeper context modeling.

\vspace{-0.6em}

\subsection{Rhetorical Role Labeling Corpora}

RRL has been explored in multiple domains using sentence-level annotation of functional discourse roles.
Starting by medical domain, \textsc{PubMed-20K-RCT}~\cite{dernoncourt-etal-2017-neural} provides a large-scale corpus of abstracts, where each sentence is labeled with a role such as \rrl{Objective}, \rrl{Methods}, or \rrl{Results}.
Similarly, \textsc{CS-Abstracts}~\cite{cohan-etal-2019-pretrained, gonçalves_2020} offers scientific abstracts with comparable rhetorical structures.

In legal NLP, research has shifted from short abstracts to full-length case documents. Corpora such as \textsc{DeepRhole}~\cite{bhattacharya2023deeprhole}, \textsc{LegalEval}~\cite{kalamkar-etal-2022-corpus}, and \textsc{LegalSeg}~\cite{nigam-etal-2025-legalseg} annotate Indian case law with roles including \rrl{Facts}, \rrl{Arguments}, and \rrl{Analysis}.
To our knowledge, no RRL corpus covers U.S. Supreme Court decisions, despite their importance for cross-jurisdictional legal applications~\cite{curry2008looking}.

\vspace{-0.6em}

\subsection{Prototype-Based Learning}

Corpus-level regularities that recur across documents are often overlooked by standard architectures. Prototype-based learning~\cite{snell2017} addresses this limitation by representing each class with a prototype, defined as the mean embedding of its support examples, and classifying new instances by measuring their proximity to these prototypes. This paradigm has achieved strong results in emotion recognition~\cite{song-etal-2022-supervised}, relation extraction~\cite{chen-etal-2023-consistent}, and named entity recognition~\cite{huang-etal-2023-pram, wu-etal-2023-mproto}, where prototypes capture class-level semantics and enable generalization under limited supervision.
\\
However, prototype-based methods remain underexplored in discourse-level classification tasks such as RRL, which could benefit from their ability to capture global semantic regularities.

\section{Enriching Discourse Representations through Semantic Prototypes}
\label{s:methodo}

\input{figures/pipeline}

This section first presents the task definition of RRL (\ssec{ss:task_def}), then outlines the backbone hierarchical architecture used in our study (\ssec{ss:backbone}). We finally introduce our global semantic prototype-based methods, Prototype-Based Regularization (\ssec{ss:PBR}) and Prototype-Conditioned Modulation (\ssec{ss:PCM}), illustrated in Figure~\ref{fig:pipeline}.

\subsection{Task Definition}
\label{ss:task_def}

Given a document $x = \{x_1, x_2, \ldots, x_m\}$ with $m$ sentences as the input, where $x_i = \{x_{i1}, x_{i2}, \ldots, x_{in}\}$ represents the $i^{\text{th}}$ sentence containing $n$ tokens, and $x_{jp}$ refers to the $p^{\text{th}}$ token in the $j^{\text{th}}$ sentence, the task of rhetorical role labeling is to predict a sequence $y = \{y_1, y_2, \ldots, y_m\}$, where $y_i$ is the rhetorical role corresponding to sentence $x_i$, and $y_i \in \mathcal{Y}$, which is the set of predefined rhetorical role labels.

\subsection{Backbone Hierarchical Architecture}
\label{ss:backbone}

All our experiments are based on the Hierarchical Sequential Labeling Network~\cite{jin-szolovits-2018-hierarchical, brack2024sequential}, the state-of-the-art RRL model designed to capture local context by modeling intra-document dependencies at multiple levels. 
Each sentence \(s_{ij}\) is first encoded with BERT~\cite{devlin-etal-2019-bert}, producing contextualized token embeddings, which are then passed through a Bi-LSTM~\cite{hochreiter1997long} and an attention-pooling mechanism~\cite{yang2016hierarchical} to obtain fixed-size sentence vectors.  
A second Bi-LSTM contextualizes these vectors with surrounding sentences, yielding enriched sentence representations, and a Conditional Random Field (CRF) layer finally predicts the optimal sequence of role labels (see Appendix~\ref{sec:architecture} for details).


\subsection{Prototype-Based Regularization}
\label{ss:PBR}

To enrich hierarchical architectures with global information, we introduce Prototype-Based Regularization (PBR), which incorporates trainable soft prototypes. These prototypes share the embedding space with sentence vectors and are optimized across documents. Rather than altering the backbone, PBR adds an auxiliary constraint that aligns each sentence embedding with its nearest prototype via a distance-based metric, steering the representation space toward corpus-level rhetorical patterns.

Following~\citet{mind_2019, Zhang_Liu_Wang_Lu_Lee_2022}, the total loss combines standard classification with two prototype-driven regularization terms: the first enforces proximity between sentences and their relevant prototypes, while the second encourages separation among prototypes to reduce redundancy in the latent space.

\vspace{-1.8em}

\begin{equation}
\mathcal{L} =
\underbrace{\mathcal{L}_{\text{task}}}_{\text{\tiny cross-entropy}} 
+ \lambda_{\text{prox}}\, \underbrace{\mathcal{L}_{\text{prox}}}_{\text{\tiny prototype proximity}} 
- \lambda_{\text{div}}\, \underbrace{\mathcal{L}_{\text{div}}}_{\text{\tiny prototype diversity}}
\label{eq:total_loss}
\end{equation}
\vspace{-0.9em}
\\
where \(\lambda_{\text{prox}}, \lambda_{\text{div}} \geq 0\) are hyperparameters controlling the contribution of each auxiliary term.

\textit{Task loss} \(\mathcal{L}_{\text{task}}\) is the standard cross-entropy computed between the model’s prediction
\(\hat{y}_{y_{ij}}\) and the gold label \(y_{ij}\) for each sentence \(s_{ij}\):

\vspace{-0.7em}

\begin{equation}
\mathcal{L}_{\text{task}}
= -\sum_{i=1}^{M} \sum_{j=1}^{N_i}
      \log \hat{y}_{y_{ij}}(s_{ij}).
\end{equation}
where $M$ denotes the number of documents in the dataset, and $N_i$ denotes the number of sentences in document $i$.

\textit{Prototype-proximity loss} \(\mathcal{L}_{\text{prox}}\) pulls every sentence embedding
\(\mathbf{h}_{ij}\) toward its nearest prototype \(P_k\) among the
\(Q\) learnable prototypes:

\vspace{-1.2em}

\begin{equation}
\mathcal{L}_{\text{prox}}
= \frac{1}{T}\;
  \sum_{i=1}^{M} \sum_{j=1}^{N_i}
  \min_{k \in \{1,\dots,Q\}}
       d\!\bigl(\mathbf{h}_{ij},\,P_k\bigr),
\end{equation}

where \(T=\sum_{i=1}^{M} N_i\) is the total number of sentences.

\textit{Prototype-diversity loss} \(\mathcal{L}_{\text{div}}\) encourages the prototypes to spread out, reducing redundancy:

\vspace{-1.2em}

\begin{equation}
\mathcal{L}_{\text{div}}
= \frac{2}{Q(Q-1)}
  \sum_{\substack{k,l \in \{1,\dots,Q\}\\k\neq l}}
  d\!\bigl(P_k,\,P_l\bigr).
\end{equation}

\subsection{Prototype-Conditioned Modulation}
\label{ss:PCM}

While PBR introduces soft alignment constraints without changing the backbone, Prototype-Conditioned Modulation (PCM) injects global representations directly into the encoding process. Prototype vectors are precomputed from the training corpus and incorporated into the hierarchical architecture through conditioning modules. These global signals modulate sentence representations during both training and inference. The procedure involves three stages: document sampling, prototype extraction, and prototype injection.

\paragraph{Document sampling}
The question is whether prototype representations should be derived from the entire training corpus or from a semantically related subset, since using all documents may introduce semantic noise and reduce prototype relevance~\cite{lai-etal-2021-learning}.
We evaluate three strategies: (1) \textit{Full Sampling}, which uses all training documents; (2) \textit{Random Sampling}, which selects a uniform subset; and (3) \textit{Supervised Sampling}, which clusters semantically similar documents using embeddings and derives prototypes per cluster\footnote{For the supervised variant, we use OpenAI’s \texttt{text-embedding-3-small}~\url{https://platform.openai.com/docs/guides/embeddings/embedding-models}, which supports sequences up to $8,192$ tokens for full-document representation. Each document is encoded and grouped via K-Means clustering~\cite{ahmed2020k}, with the optimal number of clusters selected using the Silhouette score, computed per evaluation dataset.}.

\vspace{-0.4em}

\paragraph{Prototype extraction}  
Each sentence $s_{ij}$ is embedded using a domain-specific BERT model suitable for the evaluation dataset, producing a fixed-length vector $\mathbf{h}_{ij} \in \mathbb{R}^{d}$. 
For each role $r \in \mathcal{Y}$, we compute a prototype $\mathbf{p}_r$ by averaging the embeddings of the set of sentences $\mathcal{S}_r$ annotated with $r$ in the selected document pool:

\vspace{-1.15em}

\begin{equation}
\mathbf{p}_r = \frac{1}{|S_r|} \sum_{s_{ij} \in S_r} \mathbf{h}_{ij}.
\end{equation}

\paragraph{Prototype injection}  
After computing global representations for each role, we inject them into the hierarchical architecture during both training and inference. For each sentence \(s_{ij}\), we calculate its cosine similarity with the prototype set \(\{\mathbf{p}_r\}\) and assign the closest one.
Because models are sensitive to externally injected knowledge~\cite{fu-etal-2023-revisiting}, we conducted an ablation study on five conditioning strategies inspired by prior work, with details provided in Appendix~\ref{ss:injection}.

\section{The \textsc{SCOTUS-Law} Corpus}

We introduce \textsc{SCOTUS-Law}, the first publicly available dataset of U.S. Supreme Court decisions annotated with rhetorical roles. This corpus expands the scarce resources available for RRL and provides a new benchmark for discourse analysis in the legal domain.

\subsection{Corpus Compilation \& Size}

We collected decisions from CourtListener\footnote{\url{https://www.courtlistener.com/}},
an open-access legal platform. Sampling covered three dimensions:
\textbf{(1) Temporal:} 1945--2020;
\textbf{(2) Author:} 38 justices;
\textbf{(3) Thematic:} 18 groups.
Representative cases were selected from the most prolific justices in each theme, resulting in $180$ annotated decisions with $26{,}328$ sentences, a size comparable to existing RRL datasets~\cite{kalamkar-etal-2022-corpus}.

\input{tables/stats}

\subsection{Rhetorical Scheme Description}

Our annotation scheme extends the framework of~\citet{Bonnard_2025}, which adopts the Swalesian approach~\cite{swales1990genre} to the rhetorical structure of judicial opinions. As in prior work~\cite{kalamkar-etal-2022-corpus, nigam-etal-2025-legalseg}, annotations are applied at the sentence level, but we add a further layer: each sentence is assigned a \textit{step} label that captures its function in legal reasoning and its role within the broader argumentative structure. Following~\citet{Bonnard_2025}, the scheme operates at three levels of granularity (Figure~\ref{fig:final_scheme} in Appendix).
\\
\textcolor{red}{Because of the large number of role definitions, full details are provided in Appendix~\ref{ss:annotation_scheme}.}

\vspace{.5pt}

\noindent\fbox{%
\parbox{\columnwidth}{%
\textbf{Step} = Discursive Category + Rhetorical Function + Optional Attributes
}%
}

\paragraph{\highlightcat{Discursive categories.}}
These reflect the overall structure of SCOTUS opinions and include five main categories:
\begin{itemize*}
    \item \textbf{Setting the scene}: background information and procedural history;
    \item \textbf{Analysis}: reasoning and justification of the Court’s decision;
    \item \textbf{Resolution}: the outcome or final ruling;
    \item \textbf{Sources of authority}: references to legal sources such as precedent or statutes;
    \item \textbf{Announcing}: textual elements marking structural transitions.
\end{itemize*}

\paragraph{\highlightrf{Rhetorical functions.}}
These specify the communicative role played by each segment within its discursive category.  
They include argumentative roles such as justification, evaluation, comparison, or appeal to authority (statistics report in Table~\ref{tab:scotuslaw_stats}).

\paragraph{\highlightstep{Attributes.}}
To refine the rhetorical annotation, three optional attributes can be specified:
\begin{itemize*}
    \item \textbf{Type}: the nature of the content (e.g., cited authority, recalled facts);
    \item \textbf{Author}: the speaker or source of the argument (e.g., the Court, a dissenting justice);
    \item \textbf{Target}: whether the information pertains to the current case or another referenced case.
\end{itemize*}


\subsection{Annotation Process}

The process was designed in consultation with legal experts (law professors and legal practitioners).

\paragraph{Annotator Selection and Training}

Two law students were recruited through a multi-stage selection process. Each contributed $240$ hours of annotation and received $\$1200$ in compensation. Training was provided by two legal experts to ensure a solid understanding of the rhetorical role definitions.

\paragraph{Calibration}
In the initial stages, students differed in their interpretation of roles, requiring calibration. To align their understanding with expert definitions, they annotated $18$ documents already labeled by experts. Sentences that diverged from the gold annotations were highlighted, and students revised their labels accordingly. This iterative process continued until their annotations reached expert-level agreement.

\paragraph{Adjudication}
To ensure consistency, each document was double-annotated and disagreements were resolved through adjudication. Annotators first attempted to reach consensus; unresolved cases were escalated to legal experts, who applied the guidelines and made the final decision. This process produced a coherent gold standard and refined the guidelines through feedback on ambiguous cases.

\paragraph{Annotation Quality Assessment}
Inter-annotator agreement was measured with Fleiss’ Kappa~\cite{fleiss2013statistical}, which improved from $0.67$ in the initial tests to $0.72$ after calibration, indicating good consistency. At the rhetorical function level, agreement was high for \rrl{Granting certiorari}, \rrl{Quoting}, and \rrl{Giving holding to the Court}, but lower for \rrl{Stating the Court’s Reasoning} and \rrl{Recalling}, the most ambiguous functions. These results show the challenge of separating between sentences that merely recall external sources or arguments and those where the Court formulates its own reasoning, a boundary that even experts may interpret differently.

\section{Results \& Experiments}
\input{tables/overall_results}

Table~\ref{tab:results-final} reports the results (using Macro-F1 (mF1) and Weighted-F1 (wF1)) for the SOTA hierarchical model (\ssec{ss:backbone}), our proposed prototype-based methods, and an oracle that selects the gold prototype post-hoc for reference. Experiments were conducted across legal, medical, and scientific benchmarks, including our annotated \textsc{SCOTUS-Law} corpus, and statistical significance was verified over five run evaluation.  \textcolor{blue}{Full experimental details are provided in Appendix~\ref{sec:experimental}.}



\subsection{Overall \& Fine-grained Performance}

\textbf{Do prototype-based methods consistently improve results?}
Across all legal datasets, both PBR and PCM outperform the baseline. PBR yields consistent macro-F1 gains, from $+1.5$ on \textsc{SCOTUS}\textsubscript{Category} to $+4.4$ on \textsc{SCOTUS}\textsubscript{RF}, with improvements statistically significant (\(\sigma \leq 0.3\) over five runs). PCM achieves the best results on $4$ of $7$ tasks, most notably on \textsc{SCOTUS}\textsubscript{Steps}, where scores increase from $46.70\%$ to $54.03\%$.
In comparison to Mind~\cite{t-y-s-s-etal-2024-mind}, which brings limited and task-dependent improvements, our results show that explicitly structuring representations around prototypes leads to more robust and systematic gains across datasets and annotation levels.
Regarding sampling, supervised sampling provides a slight advantage on broad labels such as \textsc{SCOTUS}\textsubscript{Category}, but this effect disappears on datasets like \textsc{LegalEval} and \textsc{DeepRhole}, where all strategies perform similarly. This may stem from two factors: (i) retrieval operates at the document level, overlooking sentence-level rhetorical similarity and sometimes producing mismatched prototypes; and (ii) legal texts follow stable rhetorical patterns, so even randomly sampled documents provide useful signals despite added noise.
\noindent
\colorbox{yellow!20}{\parbox{\dimexpr\linewidth-2\fboxsep}{
\textbf{Takeaway 1.} Global prototypes, whether used as a regularization signal (PBR) or as conditioning in the encoder (PCM), consistently improve performance beyond random variation.
}}

\textbf{Do prototypes extend benefits to minority and ambiguous roles?}
On \textsc{SCOTUS}\textsubscript{RF}, the minority role \rrl{Stating the Court’s reasoning}(less than $12.14\%$ of the data), improves with PBR from $57\%$ to $60.35\%$ ($+3.35$), demonstrating gains on long-tail labels (see Table~\ref{tab:f1-comparison-rf}). On \textsc{LegalEval}, where annotation ambiguity and subtle rhetorical distinctions are frequent~\cite{kalamkar-etal-2022-corpus}, PBR still reaches $82.5\%$. Most gains come from reducing confusions between overlapping roles such as Legal Analysis and Factual Issue Descriptions, which account for over $40\%$  of baseline errors. 
\noindent
\colorbox{yellow!20}{\parbox{\dimexpr\linewidth-2\fboxsep}{
\textbf{Takeaway 2.} Prototypes are particularly effective in challenging conditions, whether data are scarce (minority roles) or label boundaries are ambiguous (overlapping functions).
}}

\textbf{How does corpus granularity reveal the need for global cues?
} As the annotation scheme progresses from broad categories to fine-grained steps, label boundaries become harder to distinguish, such as subtypes within \rrl{Analysis}. In this setting, prototypes act as semantic anchors that stabilize representations across levels of granularity. The $+3.8$ improvement on \textsc{SCOTUS}\textsubscript{Steps} indicates that the model increasingly depends on global cues when local context is insufficient, demonstrating the value of a multi-level corpus design.

\noindent
\colorbox{yellow!20}{\parbox{\dimexpr\linewidth-2\fboxsep}{
\textbf{Takeaway 3.}
Multi-level annotation not only enriches model evaluation but also pinpoints where global prototype signals become essential.
}}

\input{tables/analysis_baseline_enrichment}

\textbf{How transferable are prototype-based methods across medical and scientific abstracts?
}
Evaluating on \textsc{PubMed} and \textsc{CS-Abstracts} allows us to test whether prototype-based methods extend beyond the legal domain, since medical and scientific abstracts also exhibit rhetorical structure but in shorter and less complex texts.
PBR improves performance on both benchmarks, confirming that structural regularization remains effective whenever discourse structure is present. In contrast, PCM brings limited gains in these settings, as prototype averaging is less informative when structural variation is low. However, oracle results—up to $99.7\%$ mF1 on \textsc{CS-Abstracts}—demonstrate the potential of PCM when relevant prototypes are retrieved, underscoring retrieval quality as the main bottleneck for cross-domain generalization.

\noindent
\colorbox{yellow!20}{\parbox{\dimexpr\linewidth-2\fboxsep}{
\textbf{Takeaway 4.} 
Structural regularization with PBR transfers robustly across domains, while PCM shows high potential but is constrained by prototype retrieval quality.
}}

\subsection{Qualitative Analysis}

\input{figures/tsne_vis}

We visualize the latent space with t-SNE (Figure~\ref{fig:tsne-proto}) to illustrate how prototypes shape sentence representations. In the baseline, clusters overlap, especially between Describing and Stating the Court’s reasoning, which often co-occur due to semantic proximity. With prototypes, these roles become more clearly separated, showing that global signals improve role distinction and that prototype quality plays a key role in structuring the latent space.

\subsection{Ablation Study}

\paragraph{PBR – hyperparameters.} 
We vary three hyperparameters on \textsc{SCOTUS}\textsubscript{RF}: the prototype count, the proximity loss \(\lambda_{\text{prox}}\), and the diversity loss \(\lambda_{\text{div}}\) (see Appendix~\ref{sec:hyper-sensitivity}). 
Performance remains stable, with slight improvements up to $16$ prototypes before plateauing. Moderate regularization (\(\lambda_{\text{prox}}=0.9\), \(\lambda_{\text{div}}=0.9\)) provides the best balance between consistency and discriminability, whereas stronger penalties slightly reduce performance by overspreading the embedding space.

\paragraph{PCM – injection strategies.} 

On legal datasets such as \textsc{SCOTUS}\textsubscript{RF} and \textsc{LegalEval}, Linear Fusion achieves the best results, improving mF1 by $+2.63$ over FiLM (see Appendix~\ref{ss:injection}). This suggests that direct concatenation aligns well with the structured rhetorical patterns of legal text. On \textsc{PubMed}, by contrast, all strategies perform similarly ($F1 > 92$), with Gated Residual Addition slightly ahead, indicating that injection is less critical for shorter and structurally simpler texts.

\section{RRL in the Age of LLMs: Performance and Efficiency Comparison}

\input{tables/llm_results}

\input{tables/expert_evaluation}

The turning point for RRL came with GPT-3.5-Turbo by OpenAI\footnote{\url{https://platform.openai.com/docs/models/gpt-3.5-turbo}},
which sparked wide evaluations of LLMs via in-context prompting. Yet,~\citet{belfathi2023harnessing} showed that despite the hype, BERT encoders~\cite{devlin-etal-2019-bert} remained superior for classification in accuracy and efficiency. Instruction fine-tuning also introduces biases, especially in multiple-choice settings where models may prefer options based on position or style rather than meaning~\cite{zheng_llm_as_judge}.

Since then, both model quality and adaptation strategies have evolved. To reassess this landscape, we fine-tuned four recent open-source models (DeepSeek-70B~\cite{guo2025deepseek}, Mistral-7B~\cite{jiang2023mistral7b}, Meta-LLaMA3-8B~\cite{dubey2024llama}, and Qwen3-8B~\cite{yang2025qwen3}) with QLoRA~\cite{dettmers2023_qlora}, using the last token as the sentence embedding, without the need for prompt engineering~\cite{wang2023improving}. 
Results in Table~\ref{tab:llms} show clear progress with fine-tuned LLMs compared to the in-context evaluations of~\citet{belfathi2023harnessing}, with Mistral-7B reaching $70.29\%$ mF1 on \textsc{SCOTUS}\textsubscript{RF}. This confirms that parameter-efficient fine-tuning can improve LLM performance for classification tasks. However, as observed in prior work~\cite{naguib-etal-2024-shot}, LLMs still tend to underperform encoder-based models when training data are sufficient, and their advantage mainly appears in low-resource or few-shot settings.


Yet, in terms of resource efficiency, our prototype-based methods (PBR and PCM) remain highly competitive, with only $110$M trainable parameters, surpassing Mistral-7B while requiring about $\sim 70\times$ fewer parameters, showing that targeted inductive biases can outperform LLMs at a fraction of the cost.

\noindent
\colorbox{yellow!20}{\parbox{\dimexpr\linewidth-2\fboxsep}{
\textbf{Takeaway 5.} 
Recent LLMs fine-tuned with QLoRA show clear progress for RRL classification, but prototype-based methods still provide a better balance between accuracy and efficiency.
}}

\section{Expert Analysis}

Following the experiments, we sought feedback from a linguistic expert with expertise in legal discourse on the accuracy of model predictions on the \texttt{\textsc{SCOTUS}\textsubscript{RF}} dataset.
To complement our quantitative results, the expert evaluation focuses on two questions: (i) whether our proposed method better resolves ambiguities between closely related rhetorical roles, and (ii) whether its gains persist across sentences of varying annotation difficulty.

\subsection{Expert Assessment of Ambiguous Rhetorical Role Pairs}

On the positive side, the expert confirmed a strong ambiguity between \rrl{Recalling} and \rrl{Stating the Court’s Reasoning}. This gray zone typically arises when a sentence both cites a source and conveys the Court’s own stance. As shown in Table~\ref{tab:expert-eval}, our PCM approach reduced errors for this pair by $19.75\%$ compared to the baseline. A similar improvement was observed for the pair \rrl{Recalling} and \rrl{Quoting}, where ambiguity emerges when a sentence combines quotation with commentary or relies on partial citations as supporting evidence.

However, the expert also emphasized that certain roles remain difficult to classify. In particular, \rrl{Accepting} and \rrl{Rejecting} arguments are frequently mislabeled as \rrl{Recalling} when sources are cited, even in the presence of explicit evaluative markers such as \textit{“we agree”}, \textit{“was correct to point out”}, or \textit{“we disapprove”}.

\input{tables/hardness}

\subsection{Expert Assessment Across Annotation Difficulty Levels}

We conducted an additional evaluation with a legal expert. We sampled $150$ segments from \textsc{SCOTUS}\textsubscript{RF} and asked the expert to assess their annotation difficulty using a Likert scale (1 = easy, 4 = difficult). We then measured the accuracy of our best-performing system (PCM) across these difficulty levels.

The results presented in Table~\ref{tab:difficulty-macro} show that PCM achieves perfect performance on clearly defined rhetorical roles, remains robust on moderately easy cases, and retains meaningful predictive ability on difficult segments where even experts report substantial ambiguity. As difficulty increases, errors tend to correspond to well-known confusions between closely related rhetorical functions (e.g., \rrl{Recalling} vs. \rrl{Stating the Court’s Reasoning}), thereby reinforcing the insights drawn from our qualitative analysis.

\noindent
\colorbox{yellow!20}{\parbox{\dimexpr\linewidth-2\fboxsep}{
\textbf{Takeaway 6.} 
Expert analysis confirms that prototypes reduce key ambiguities, but some functions remain challenging, highlighting the need for future model improvements.
}}

\section{Discussion}

Existing approaches to RRL, particularly on LegalEval~\cite{kalamkar-etal-2022-corpus, belfathi-selective-2025}, are robust but still fail to separate roles that are semantically close. By integrating local context with global prototypes, our methods reduce such ambiguities, as shown both by consistent performance gains and by expert analysis on our new dataset.

Differences across corpora should be interpreted cautiously: macro-F1 variations reflect not only model quality, but also label inventories, granularity, class imbalance, and document style. In particular, \textsc{LegalEval} uses an annotation scheme that differs from \textsc{SCOTUS}\textsubscript{RF}, so the two are not strictly comparable in terms of absolute scores. A more meaningful cross-corpus comparison is therefore the relative improvement over a common baseline, which remains stable across datasets and suggests that prototype-based structuring is broadly beneficial. Lower performance on \textsc{DeepRhole}, despite its smaller label set, is also expected: fewer roles can correspond to broader and more heterogeneous classes, making boundaries harder to learn. This highlights that task difficulty is not determined by the number of labels alone, but by their separability and consistency in the corpus.

Our corpus centers on long judicial decisions, evaluated with five-fold cross-validation on $18$ test cases ($2{,}481$ sentences). The low variance across splits ($\pm0.3$ mF1) confirms that improvements are systematic rather than data-dependent.

The comparison of PBR and PCM highlights complementary strengths. PBR acts as a lightweight regularizer, training $20$--$25\%$ faster and using $30$--$40\%$ less GPU memory, which makes it well suited to resource-limited settings. PCM is more computationally demanding due to prototype precomputation and conditioning, but it delivers stronger performance gains, especially for fine-grained roles.

Finally, regarding efficiency,~\citet{belfathi2023harnessing} underscored the high cost of LLMs for RRL. Our results suggest that prototype-based methods achieve a more favorable accuracy–efficiency trade-off, outperforming recent LLM-based systems under constrained resources.

\section{Conclusion}

This work shows that combining local context with global semantic prototypes significantly improves RRL, particularly for underrepresented roles. By introducing two methods—Prototype-Based Regularization (PBR) and Prototype-Conditioned Modulation (PCM)—we show that global signals can be effectively injected into hierarchical architectures to provide more semantically coherent representations. Beyond model performance, we contribute \textsc{SCOTUS-Law}, the first U.S. Supreme Court dataset annotated at three rhetorical levels. This resource enables more granular evaluation and promotes research on the legal NLP field. Our findings also indicate that prototype-based methods offer a more favorable accuracy–efficiency trade-off than fine-tuned LLMs. 
Future work should prioritize extending semantic prototyping to multilingual and cross-domain RRL, where generalization is even more demanding.


\section{Limitations}

Although the proposed methods improve RRL performance, several limitations should be acknowledged to guide future improvements:

\begin{itemize*}
    \item The current problem formulation assigns a single rhetorical label to each sentence in a multi-class classification task. While this simplifies annotation and modeling, it may not account for the semantic complexity of long or compound sentences that express multiple rhetorical functions. Reformulating the task as multi-label classification could better reflect such cases.

    \item The approach operates at the sentence level. Segmenting at the phrase or clause level, and modeling rhetorical dependencies between segments, could lead to more fine-grained analysis.

    \item The study focuses exclusively on English corpora. Extending semantic prototyping to multilingual RRL raises challenges related to alignment, label transfer, and prototype sharing across languages with different rhetorical conventions.

\end{itemize*}

\section{Ethical considerations}

This work proposes new methods and experiments aimed at advancing research in rhetorical role labeling, a foundational task in legal document processing. All experiments were conducted on publicly available datasets, including our introduced datasets. While these documents are not anonymized and may contain real names of involved parties, they are official court records released for public access.
We do not anticipate any harm arising from our use of these datasets. Our research is intended to support the development of transparent and responsible AI tools for legal professionals. By improving the automation of rhetorical role labeling, we aim to facilitate legal text analysis and contribute positively to the broader goals of legal NLP.

\section{Acknowledgments}

This work was granted access to the HPC resources of IDRIS under the allocations 2023-AD011014882 and 2023-AD011014767, provided by GENCI.

This research was funded, in whole or in part, by l’Agence Nationale de la Recherche (ANR), project ANR-22-CE38-0004.


\bibliography{custom}

\bibliographystyle{acl_natbib}

\clearpage

\appendix

\section{Annotation Scheme}
\label{ss:annotation_scheme}

Figure~\ref{fig:final_scheme} presents the complete annotation scheme, whose main components are detailed below.

\subsection{\highlightcat{Discursive Categories}}
\label{sec:categories}

The first level of our annotation schema defines five high-level rhetorical categories that segment each decision into major structural blocks. Below, we provide a brief description of each one:

\paragraph{Setting the scene.}  
This category includes introductory paragraphs that present the case to the reader. Typical content includes information about the nature of the parties involved, their claims, the material facts of the case, the legal issue under examination, and the procedural history that brought the case before the Supreme Court.

\paragraph{Analysis.}  
This category corresponds to the argumentative core of the decision. It usually follows the introductory section and precedes the final ruling. The content is primarily argumentative and captures the Court's reasoning in response to the parties’ claims, justifying the interpretation and application of legal principles.

\paragraph{Resolution.}  
This section contains the resolution of the legal issue, typically expressed through the final ruling issued by the majority opinion. While the announcement of the judgment is obligatory, it may also include instructions for lower courts or comments on the societal impact of the decision.

\paragraph{Sources of authority.}  
This category gathers all explicit mentions of legal sources, whether written (e.g., case law, statutes, constitutional texts) or unwritten (e.g., doctrines or principles). Although such references appear throughout the decision, some judges explicitly dedicate specific portions of their opinion to outlining the sources that will later support their legal reasoning.  
\textit{Note:} when a source is invoked directly within the reasoning process, it is annotated under the \textit{Analysis} category rather than \textit{Authoritative Sources}.

\paragraph{Announcing.}  
This category includes structurally functional sentences that serve as rhetorical transitions. These statements do not carry substantive content themselves but signal the upcoming development of a new rhetorical step from one of the four other categories.

\subsection{\highlightrf{Rhetorical Functions}}
\label{sec:rhetorical-functions}

At the second level of annotation, we define thirteen rhetorical functions that capture the specific communicative intent of each sentence in the decision. 

\paragraph{Granting certiorari.}  
Assigned to sentences where the Court explicitly signals that it has agreed to review the case. These statements typically appear near the end of the factual and procedural summary, often preceding the articulation of the legal questions. Example: “We granted certiorari.”

\paragraph{Presenting jurisdiction.}  
Covers sentences that neutrally present elements of the case background. This function includes an attribute \texttt{Type} with five possible values: \textit{Legal Issue}, \textit{Facts of the Case}, \textit{Other Procedural Elements}, \textit{Arguments and Claims}, or \textit{Broader Context}.

\paragraph{Quoting.}  
Used for references to legal sources. The annotation includes a \texttt{Type} indicating the nature of the source: \textit{Court Decision}, \textit{Primary Source}, or \textit{Secondary Source}.

\paragraph{Describing.}  
Applied to paraphrases of legal sources, whether primary, secondary, or unwritten. The associated \texttt{Type} indicates the source category: \textit{Primary Source}, \textit{Secondary Source}, or \textit{Unwritten Source of Authority}.

\paragraph{Citing.}  
Used for direct quotations that include complete sentences or longer excerpts from legal sources. Types are the same as for \textit{Quoting}.

\paragraph{Recalling.}  
Captures sentences that refer back to previously mentioned legal sources, or that introduce sources in a way that supports the Court’s reasoning. These recalls often include an interpretive dimension, contributing to argumentative development.

\paragraph{Accepting arguments/a reasoning.}  
Marks agreement with a previously stated argument or reasoning, either from a party or another court.

\paragraph{Rejecting arguments/a reasoning.}  
Indicates disagreement or refutation of a prior argument or line of reasoning, particularly when opposing the view of another court.

\paragraph{Stating the Court’s reasoning.}  
Assigned to all reasoning sentences that do not fall under more specific categories. This includes hypothetical reasoning, such as evaluating consequences of alternative outcomes.

\paragraph{Giving instructions to competent courts.}  
Covers sentences in which the Court instructs lower courts or other legal bodies to act in accordance with the decision or to reconsider aspects of the case.

\paragraph{Giving the holding of the Court.}  
Applies to sentences stating the legal conclusion reached by the Court (the holding), based on the material facts, including the final judgment.

\paragraph{Evaluating the impact of the decision.}  
Used when the Court explicitly reflects on the consequences of its decision, either institutionally or societally.

\paragraph{Announcing.}  
Marks structurally functional sentences that introduce an upcoming element of the decision or name the judge who authored the opinion.

\subsection{\highlightstep{Attributes}}
\label{sec:attributes}

To enrich the rhetorical annotation while keeping the core label space concise, we introduce a small set of optional attributes. These attributes are designed to add interpretive nuance without changing the primary function assigned to a sentence. They are used selectively with certain rhetorical functions, such as \textit{Recalling}, \textit{Describing}, or \textit{Presenting jurisdiction}.

\begin{itemize}
    \item \textbf{Type} — indicates the nature of the content referenced or discussed (e.g., legal source, factual detail, procedural element);
    \item \textbf{Author} — specifies who is the originator of the argument or point of view (e.g., the Court, a party, or a dissenting opinion);
    \item \textbf{Target} — identifies whether the information concerns the case under review or refers to another precedent.
\end{itemize}

These attributes are optional but help clarify rhetorical intent, especially in ambiguous or multi-voiced legal discourse.

\section{Corpus Statistics}

This section presents key statistics and descriptive analyses of the annotated \textsc{SCOTUS-Law} corpus. Figure~\ref{fig:meta-stats} illustrates the topical, temporal, and authorial diversity of the dataset, showing the wide range of judicial opinions included. Figure~\ref{fig:relative_pos} depicts the structural flow of rhetorical functions within legal cases, highlighting their relative positions throughout a decision. Finally, Table~\ref{tab:codingscheme_and_number_of_annotations} reports the number and proportion of instances across the three annotation levels—\textit{category}, \textit{rhetorical function}, and \textit{step}—providing an overview of the corpus composition.

\input{figures/meta-data-distr}
\input{figures/position_rhetorical_functions}
\input{tables/stats-corpus}
\input{figures/tagset}

\section{Hierarchical Architecture Details}
\label{sec:architecture}

\input{figures/hierarchical_architecture}

All of our experiments are built on the state-of-the-art hierarchical architecture~\cite{brack2024sequential}. 
Initially, each sentence \(s_{ij}\) is encoded independently with a BERT model~\cite{devlin-etal-2019-bert}, producing a sequence of contextual token embeddings  
\(\mathbf{h}_{ij}=\{\mathbf{h}_{ij1},\mathbf{h}_{ij2},\ldots,\mathbf{h}_{ijT_{ij}}\}\).  
These vectors are passed through a Bi-LSTM layer~\cite{hochreiter1997long}, followed by an attention-pooling layer~\cite{yang2016hierarchical}, to yield sentence representations
\(\mathbf{v}_{ij}\).

\begin{equation}
  \mathbf{u}_{ijt}=\tanh\!\bigl(W_w\mathbf{h}_{ijt}+\mathbf{b}_w\bigr)
\end{equation}

\begin{equation}
  \alpha_{ijt}=
    \frac{\exp\!\bigl(\mathbf{u}_{ijt}^{\top}\mathbf{u}_w\bigr)}
         {\sum_{t'}\exp\!\bigl(\mathbf{u}_{ijt'}^{\top}\mathbf{u}_w\bigr)}
  \quad\&\quad
  \mathbf{v}_{ij}= \sum_{t=1}^{T_{ij}}\alpha_{ijt}\,\mathbf{h}_{ijt}
\end{equation}

Here, \(W_w\), \(\mathbf{b}_w\), and \(\mathbf{u}_w\) are trainable parameters.  
The sentence representations \(\mathbf{v}_{ij}\) are then passed through a second Bi-LSTM to obtain contextualised embeddings  
\(\mathbf{c}_{ij}\) that capture information from neighbouring sentences.  
Finally, the contextual vectors \(\mathbf{c}_{ij}\) are fed to a Conditional Random Field layer, which predicts the optimal sequence of labels.

\section{Experimental Setup}
\label{sec:experimental}

For all datasets, data are split at the document level to prevent any cross-split leakage, ensuring a clean separation between training, development, and test sets.

\subsection{Evaluation Datasets}
\label{ss:evaluation_dataset}

\input{tables/datasets_details}

In addition to evaluating our models on the proposed \textsc{Scotus-Law} corpus, we conduct experiments on several established RRL benchmarks across the legal, medical, and scientific domains.

\vspace{0.5em}
\noindent
\textbf{LegalEval}~\cite{kalamkar-etal-2022-corpus} consists of judgments from the Indian Supreme Court, High Court, and District Courts. It provides public training and validation splits with 184 and 30 documents, respectively, totaling 31{,}865 sentences (average of 115 per document), annotated with 13 rhetorical role labels. Due to the absence of a public test set, we train on the official training split and evaluate on the provided validation set.

\vspace{0.5em}
\noindent
\textbf{DeepRhole}~\cite{bhattacharya2023deeprhole} includes 50 judgments from the Indian Supreme Court across five legal domains, annotated with 7 rhetorical roles. It comprises 9{,}380 sentences (average of 188 per document). We follow an 80/10/10 split at the document level for train/validation/test.

\vspace{0.5em}
\noindent
\textbf{PubMed}~\cite{dernoncourt-lee-2017-pubmed} contains 20{,}000 structured medical abstracts from randomized controlled trials. Sentences are automatically labeled by authors into five rhetorical roles: \textit{Background}, \textit{Objective}, \textit{Methods}, \textit{Results}, and \textit{Conclusions}.

\vspace{0.5em}
\noindent
\textbf{CS-Abstracts}~\cite{gonçalves_2020} includes 654 abstracts from computer science literature, annotated via crowdsourcing into the same five rhetorical roles as PubMed. It is currently the most recent dataset for scientific rhetorical structure classification.

\subsection{PBR Hyperparameters}

Following~\citet{chen_2019}, we use cosine similarity to compute distances \(d\) between sentence embeddings and prototypes.  
To control the granularity of the soft prototype space, we vary \(Q \in \{2, 4, 8, 16, 32, 64\}\) , as in~\citet{yang2018robust, SOURATI2023110418}.  
The auxiliary loss weights \(\lambda_{\text{prox}}\) and \(\lambda_{\text{div}}\) are tested over \(\{0, 0.9, 10\}\), where \(\lambda = 0\) disables the constraint, 0.9 is a balanced setting from~\citet{das-etal-2022-prototex}, and 10 enforces strong regularization.

\subsection{PCM Hyperparameters}

In supervised sampling, documents are clustered by semantic similarity.
The number of clusters is tuned on the development set using the silhouette score over the range \([1, 10]\).  
For prototype extraction, we use \texttt{Legal-BERT-uncased}~\cite{chalkidis-etal-2020-legal} for legal data, and \texttt{SciBERT-uncased}~\cite{beltagy-etal-2019-scibert} for medical and scientific domains.

\subsection{Implementation Details}
\label{sec:implementation}

We follow the hyperparameters for the baseline as described in \citet{brack2024sequential}. We use the \textsc{BERT}-\textit{base} model to obtain the token encodings. We employ a dropout of 0.5, a maximum sequence length of 128, an LSTM dimension of 768, and an attention context dimension of 200. 
We perform a grid search over learning rates \{1e-5, 3e-5, 5e-5, 1e-4, 3e-4\} for 40 epochs, using the Adam optimizer \citep{kingma2014adam}.

\section{Sensitivity to PBR Hyperparameters}
\label{sec:hyper-sensitivity}

We evaluate PBR sensitivity on \textsc{SCOTUS}\textsubscript{RF}, focusing on three components: (1) the number of soft prototypes, (2) the proximity loss weight \(\lambda_{\text{prox}}\), and (3) the diversity loss weight \(\lambda_{\text{div}}\), as shown in Figure~\ref{fig:rf-proto-lambdas-final}.

\paragraph{Prototype count.}  
Performance is stable across values, with a slight improvement up to $16$ prototypes. Beyond that, gains plateau, suggesting that few prototypes suffice to capture key rhetorical patterns, while higher counts may introduce redundancy.

\paragraph{Proximity loss \(\lambda_{\text{prox}}\).}  
A moderate value (\(\lambda_{\text{prox}} = 0.9\)) yields the best results, supporting the idea that proximity improves role consistency. Higher pressure (\(\lambda_{\text{prox}} = 10.0\)) degrades performance, likely due to overcompression of the embedding space.

\paragraph{Diversity loss \(\lambda_{\text{div}}\).}  
An intermediate value \(\lambda_{\text{div}} = 0.9\) also performs best. It encourages separation among prototypes, improving class discriminability. Stronger regularization (\(\lambda_{\text{div}} = 10.0\)) slightly hurts performance, possibly by pushing prototypes too far from the data manifold.

\section{Prototype Injection Strategies}
\label{ss:injection}

\input{tables/injection}
\input{figures/protototype_sensitivity}

We experiment with several strategies to inject global prototype representations into sentence encoders. Each method varies in the degree of control, parametrization, and how the prototype signal is merged with the original sentence representation. We describe below the five main approaches studied in our work.

\paragraph{Linear Fusion}~\cite{bu-etal-2023-segment} 
This method concatenates the sentence and its corresponding prototype vector, followed by a linear projection layer to recover the original embedding dimension. While simple and fully parametric, this technique may dilute the prototype signal due to compression.

\paragraph{Conditional Layer Normalization (CLN)}~\cite{lee-etal-2021-enhancing} 
The sentence is first normalized (zero mean, unit variance), and the prototype generates two vectors $\gamma$ (gain) and $\beta$ (bias) that re-scale and shift each dimension of the sentence embedding. This conditioning allows for fine-grained recalibration informed by prototype semantics.

\paragraph{Gated Residual Addition}~\cite{tsur-tulpan-2023-deeper} 
The original sentence embedding is preserved, and a prototype-based residual is added with a learned gate vector $g \in [0,1]^d$ that controls per-dimension contribution. If $g$ closes, the model reverts to the baseline representation; if it opens, the prototype is effectively injected.

\paragraph{Feature-wise Linear Modulation (FiLM)}~\cite{ahrens-etal-2023-visually} 
FiLM extends CLN by directly applying the prototype-derived $\gamma$ and $\beta$ vectors to modulate the sentence features ($\gamma \odot x + \beta$), without requiring prior normalization. This method is more flexible but less controlled than CLN, enabling adaptive influence of the prototype on the sentence.

\paragraph{Cross-Attention Fusion}~\cite{zhang-etal-2024-coarse} 
Here, the sentence acts as a query vector, attending to the prototype treated as key/value. Attention weights select relevant components from the prototype to be added to the sentence. This dynamic fusion allows for sentence-specific contextualization, adapting the contribution of the prototype to the input.

\vspace{0.3em}
Each mechanism provides a different trade-off between interpretability, efficiency, and contextual adaptation. The ablation study in Table~\ref{tab:pcm-injection-wf1} shows that no method is universally optimal, and that effectiveness often depends on the nature of the data and task.

\end{document}

%% file: figures/example.tex
\begin{figure*}[ht]
  \centering
  \includegraphics[width=\textwidth]{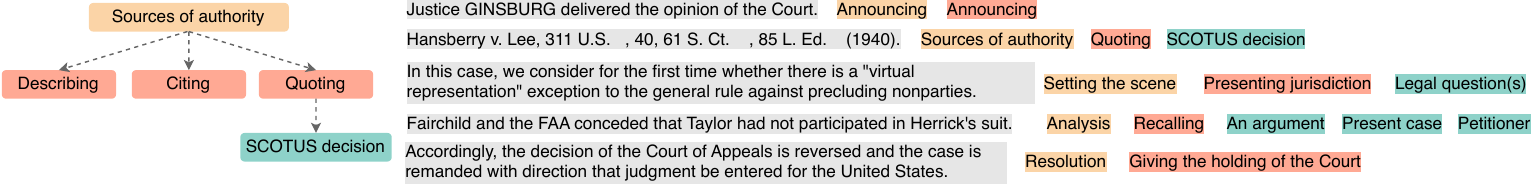}
  \caption{
Example of a document segments in \textsc{SCOTUS-Law}, annotated with \highlightc{discursive categories}, \highlightr{rhetorical functions}, and \highlightat{attributes}, which together form the step annotation (full hierarchy in Figure~\ref{fig:final_scheme}).
}
  \label{fig:exemple}
  \vspace{-0.7em}

\end{figure*}

%% file: figures/pipeline.tex
\begin{figure*}[ht]
  \centering
  \includegraphics[width=\textwidth]{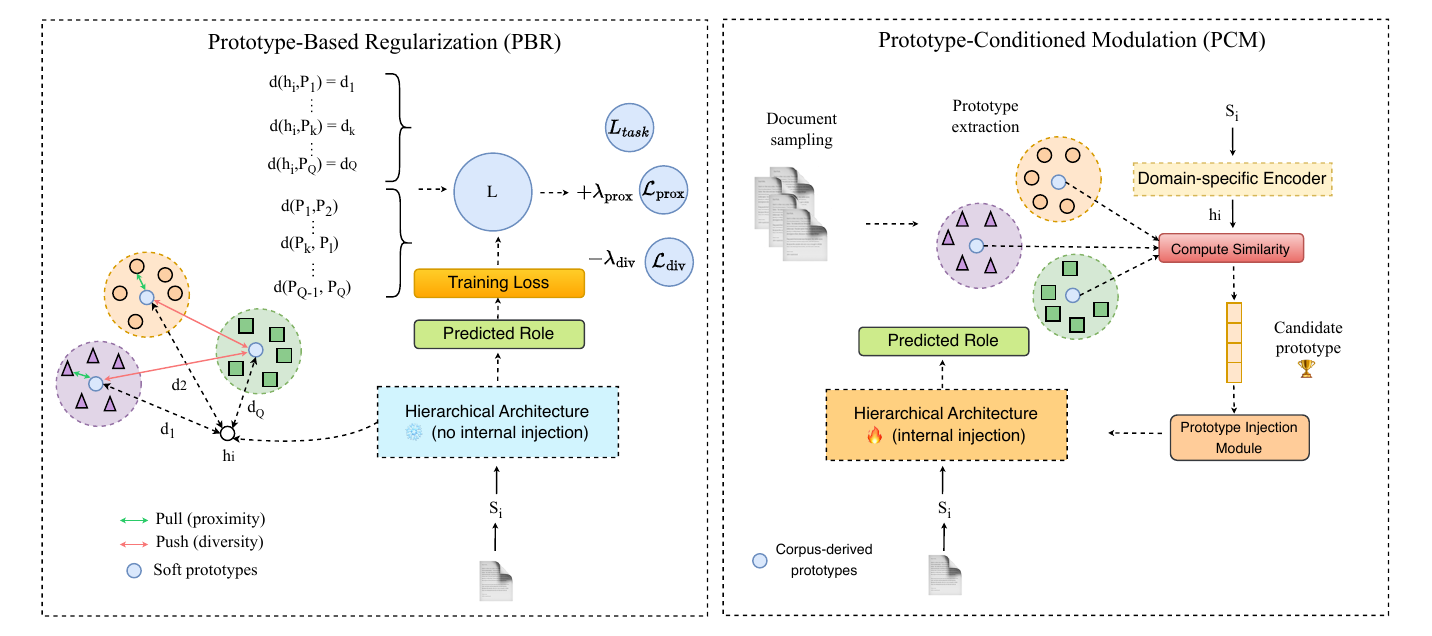}
  \caption{Illustration of our methods for injecting global representations into hierarchical architectures.
PBR (left) learns soft prototypes jointly with the model to regularize the latent space.
PCM (right) dynamically injects precomputed role prototypes during encoding via modulation mechanisms.}
  \label{fig:pipeline}
  \vspace{-1.0em}

\end{figure*}

%% file: tables/stats.tex
\begin{table}[t!]
\centering
\small
\arrayrulecolor{gray!30}
\rowcolors{2}{white}{gray!08}
\setlength{\tabcolsep}{6pt}
\resizebox{\linewidth}{!}{%
\begin{tabular}{lccc}
\toprule
\multicolumn{4}{c}{\textbf{Corpus-level statistics}} \\
\midrule
\textbf{Statistic} & \textbf{Train} & \textbf{Dev} & \textbf{Test} \\
\midrule
\# Documents               & 144     & 18    & 18    \\
Total \# Sentences        & 21,396  & 2,450 & 2,481 \\
Avg. \# Sentences / Doc   & 148.58  & 136.11& 137.83 \\
Avg. \# Tokens / Sentence & 22.95   & 21.43 & 22.15 \\
\midrule
\multicolumn{4}{c}{\textbf{Sentence distribution by rhetorical function}} \\
\midrule
\textbf{Label} & \multicolumn{3}{c}{\textbf{Total (percentage)}} \\
\midrule
Recalling                               & \multicolumn{3}{c}{8,119 \;\;(30.8\%)} \\
Quoting                                 & \multicolumn{3}{c}{6,441 \;\;(24.5\%)} \\
Presenting jurisdiction                 & \multicolumn{3}{c}{4,941 \;\;(18.8\%)} \\
Stating the Court’s reasoning           & \multicolumn{3}{c}{3,198 \;\;(12.1\%)} \\
Describing                              & \multicolumn{3}{c}{\hspace{1em}955 \;\;(3.6\%)} \\
Giving the holding of the Court         & \multicolumn{3}{c}{\hspace{1em}760 \;\;(2.9\%)} \\
Citing                                  & \multicolumn{3}{c}{\hspace{1em}644 \;\;(2.4\%)} \\
Rejecting arguments/a reasoning         & \multicolumn{3}{c}{\hspace{1em}490 \;\;(1.9\%)} \\
Announcing                              & \multicolumn{3}{c}{\hspace{1em}344 \;\;(1.3\%)} \\
Granting certiorari                     & \multicolumn{3}{c}{\hspace{1em}182 \;\;(0.7\%)} \\
Giving instructions to competent courts & \multicolumn{3}{c}{\hspace{1em}105 \;\;(0.4\%)} \\
Accepting arguments/a reasoning         & \multicolumn{3}{c}{\hspace{1em}103 \;\;(0.4\%)} \\
Evaluating the impact of the decision   & \multicolumn{3}{c}{\hspace{1em}\;45 \;\;(0.2\%)} \\
\bottomrule
\end{tabular}%
}
\caption{Descriptive statistics for the \textsc{SCOTUS-Law} dataset at the rhetorical function level.}

\label{tab:scotuslaw_stats}
  \vspace{-1.4em}

\end{table}

%% file: tables/overall_results.tex
\begin{table*}[ht]
\centering
\small    
\resizebox{\linewidth}{!}{  
\begin{tabular}{>{\bfseries}lcccccccccccccccc}
\toprule
& \multicolumn{10}{c}{\includegraphics[height=1.2em]{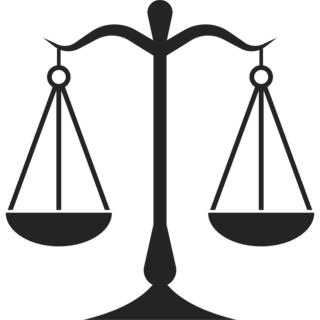}~~\textbf{Legal}}  & \multicolumn{2}{c}{\includegraphics[height=1.2em]{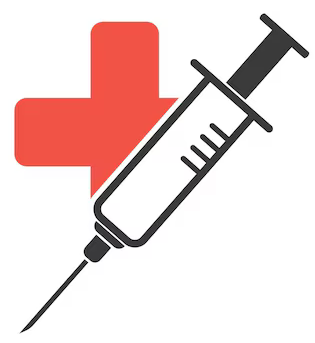}~~\textbf{Medical}}  & \multicolumn{2}{c}{\includegraphics[height=1.2em]{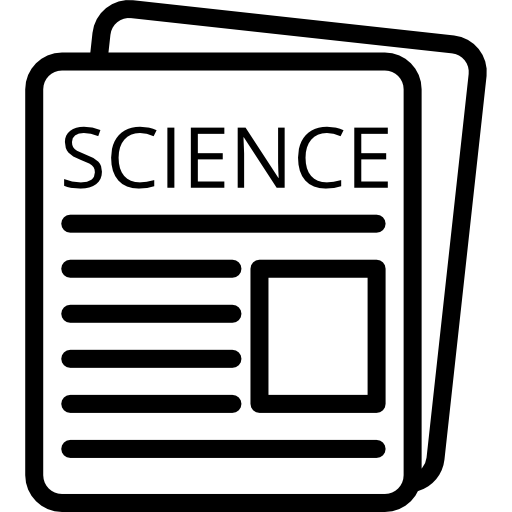}~~\textbf{Scientific}} \\
\cmidrule(lr){2-11} \cmidrule(lr){12-13} \cmidrule(lr){14-15}
& \multicolumn{2}{c}{\textsc{Scotus}\textsubscript{Category}} 
& \multicolumn{2}{c}{\textsc{Scotus}\textsubscript{RF}} 
& \multicolumn{2}{c}{\textsc{Scotus}\textsubscript{Steps}} 
& \multicolumn{2}{c}{\textsc{LegalEval}} 
& \multicolumn{2}{c}{\textsc{DeepRhole}}
& \multicolumn{2}{c}{\textsc{PubMed}} 
& \multicolumn{2}{c}{\textsc{CS-Abstracts}} \\
\cmidrule(lr){2-3} \cmidrule(lr){4-5} \cmidrule(lr){6-7}
\cmidrule(lr){8-9} \cmidrule(lr){10-11} \cmidrule(lr){12-13}
\cmidrule(lr){14-15}
& mF1 & wF1 
& mF1 & wF1 
& mF1 & wF1 
& mF1 & wF1 
& mF1 & wF1 
& mF1 & wF1 
& mF1 & wF1 \\
\midrule

 Baseline (HSLN)
& 82.22  & 88.35 & 61.36  & 78.81  & 46.70 & 63.21 & 78.82 & 90.94 & 44.24 & 50.51 & 87.01 & 91.09 & 68.55 & 75.08 \\

 Mind~\citet{t-y-s-s-etal-2024-mind}
& 83.46  & 89.20 & 62.67  & 79.07  & 45.24 & 62.78 & 79.80 & 91.25 & 45.30 & 50.93 & 87.67 & 91.86 & 69.19 & 76.91 \\

\midrule[0.2pt]

 PBR  
  & 83.69\sddag & 89.75\sddag & 65.75\sddag & 80.31\sddag & 50.48\sddag & 65.73\sddag & 82.50\sddag & 93.17\sddag & 44.96\sdag & 51.11\sdag & 88.86\sddag & 92.91\sddag & 71.10\sddag &  78.09\sddag \\

\midrule[0.2pt]
 PCM (Full Sampling)  
  & 83.96\sddag & 89.80\sddag & 67.53\sddag & 80.64\sddag & 54.03\sddag & 67.54\sddag & 81.41\sddag &  91.21 & 47.13\sddag & 55.54\sddag & 87.19 & 91.89 &  69.84 & 76.66 \\
 PCM (Random Sampling)  
  & 83.93\sddag &  89.70\sddag & 67.24\sddag & 80.66\sddag & 54.62\sddag & 67.55\sddag & 81.83\sddag & 91.57 & 47.30\sddag &  53.90\sddag & 87.24 & 91.94  &  69.12 & 76.30\sdag   \\
 PCM (Supervised Sampling)  
  & 84.13\sddag & 89.75\sddag & 67.45\sddag & 80.92\sddag & 54.40\sddag & 67.79\sddag & 80.77\sddag & 91.00 & 45.92\sddag & 53.86\sddag & 87.42 & 92.06\sdag &  68.69 & 75.46 \\
\midrule[0.2pt]

 Gold Prototypes  &
 85.20 & 90.02 & 68.86 & 81.11 & 56.20 & 69.86 & 91.71 &  99.57 & 47.90 & 56.02 & 100.0 &  100.0 &  99.66 & 99.84   \\

\bottomrule
\end{tabular}
}
\caption{
Macro-F1 and Weighted-F1 scores across domains for the baseline, PBR, and PCM (with different sampling strategies). 
An gold prototypes experiment is also included, selecting the optimal prototype post-hoc for each sentence. \sdag~and \sddag~indicate statistical significance over the baseline at 
\(0.05\) and \(0.01\), respectively.
}

\label{tab:results-final}
  \vspace{-1.3em}

\end{table*}

%% file: tables/analysis_baseline_enrichment.tex
\begin{table}[t!]
\centering
\small
\setlength{\tabcolsep}{4pt}
\renewcommand{\arraystretch}{1.08}
\resizebox{\columnwidth}{!}{
\begin{tabular}{l S[table-format=2.1] S[table-format=3.2] S[table-format=3.2] l}
\toprule
\textbf{Rhetorical Function} & \textbf{\%} & \textbf{Baseline} & \textbf{+PCM} & \textbf{\(\Delta\)} \\
\midrule
Accepting arguments/a reasoning         & 0.4  & 15.40 & 57.15 & \pos{41.75} \\
Announcing                              & 1.3  & 68.98 & 76.93 & \pos{7.95}  \\
Citing                                  & 2.4  & 85.99 & 89.92 & \pos{3.93}  \\
Describing                              & 3.6  & 61.04 & 61.41 & \pos{0.37}  \\
Evaluating the impact of the decision   & 0.2  &  0.00 &  0.00 & 0.00        \\
Giving instructions to competent courts & 0.4  & 52.18 & 56.01 & \pos{3.83}  \\
Giving the holding of the Court         & 2.9  & 74.63 & 81.61 & \pos{6.98}  \\
Granting certiorari                     & 0.7  & 97.30 &100.00 & \pos{2.70}  \\
Presenting jurisdiction                 &18.8  & 86.64 & 88.65 & \pos{2.01}  \\
Quoting                                 &24.5  & 97.79 & 98.13 & \pos{0.34}  \\
Recalling                               &30.8  & 77.38 & 79.04 & \pos{1.66}  \\
Rejecting arguments/a reasoning         & 1.9  & 40.52 & 35.91 & \negat{4.61}\\
Stating the Court’s reasoning           &12.1  & 57.00 & 60.35 & \pos{3.35}  \\
\midrule
\textbf{Macro-F1} & {} & 62.69 & 68.09 &
\cellcolor{green!10}\textbf{\textcolor{green!50!black}{\(\uparrow\)~5.40}} \\
\bottomrule
\end{tabular}
}
\caption{
Role-wise F1 on \textsc{SCOTUS}\textsubscript{RF}.
The \% column indicates the proportion of each rhetorical function in the corpus.
}
\label{tab:f1-comparison-rf}
\vspace{-1.8em}
\end{table}

%% file: figures/tsne_vis.tex
\begin{figure}[t]
\centering
\includegraphics[width=0.8\linewidth]{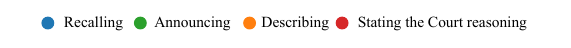}

\setlength{\tabcolsep}{0pt}  
\renewcommand{\arraystretch}{0.9}  

\begin{tabular}{ccc}
\includegraphics[width=0.325\linewidth]{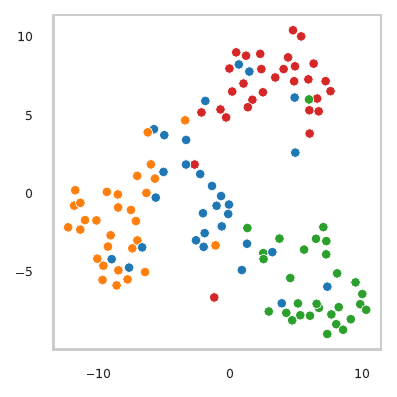} &
\includegraphics[width=0.325\linewidth]{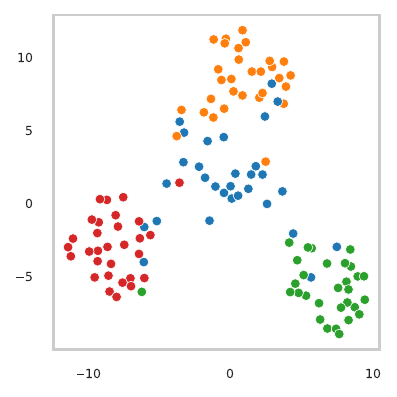} &
\includegraphics[width=0.325\linewidth]{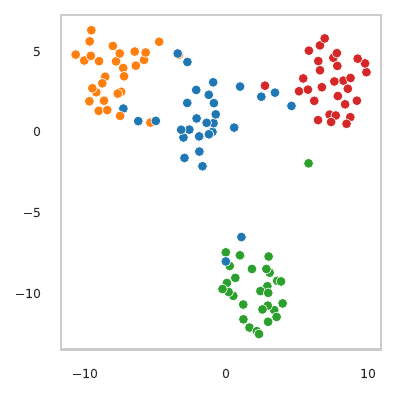} \\
\scriptsize (a) Baseline & 
\scriptsize (b) + PBR & 
\scriptsize (c) + PCM
\end{tabular}

\vspace{-0.5em}
\caption{
t-SNE projection of sentence embeddings under baseline, PBR, and PCM.
}
\label{fig:tsne-proto}
\vspace{-1.6em}

\end{figure}

%% file: tables/llm_results.tex
\begin{table}[t]
\centering
\small
\arrayrulecolor{gray!30}
\rowcolors{2}{white}{gray!08}
\resizebox{\linewidth}{!}{
\begin{tabular}{>{\bfseries}lcccccc}
\toprule
& \multicolumn{2}{c}{\includegraphics[height=1.2em]{images/legal.png}~~\textbf{Legal}} 
& \multicolumn{2}{c}{\includegraphics[height=1.2em]{images/medical.png}~~\textbf{Medical}} 
& \multicolumn{2}{c}{\includegraphics[height=1.2em]{images/science.png}~~\textbf{Scientific}} \\
\cmidrule(lr){2-3} \cmidrule(lr){4-5} \cmidrule(lr){6-7}
 & \multicolumn{2}{c}{\textsc{SCOTUS}\textsubscript{RF}} & \multicolumn{2}{c}{\textsc{PubMed}} & \multicolumn{2}{c}{\textsc{CS-Abstracts}} \\
\cmidrule(lr){2-3} \cmidrule(lr){4-5} \cmidrule(lr){6-7}
 & mF1 & wF1 & mF1 & wF1 & mF1 & wF1 \\
\midrule
\rowcolor{gray!15}
\includegraphics[height=1.2em]{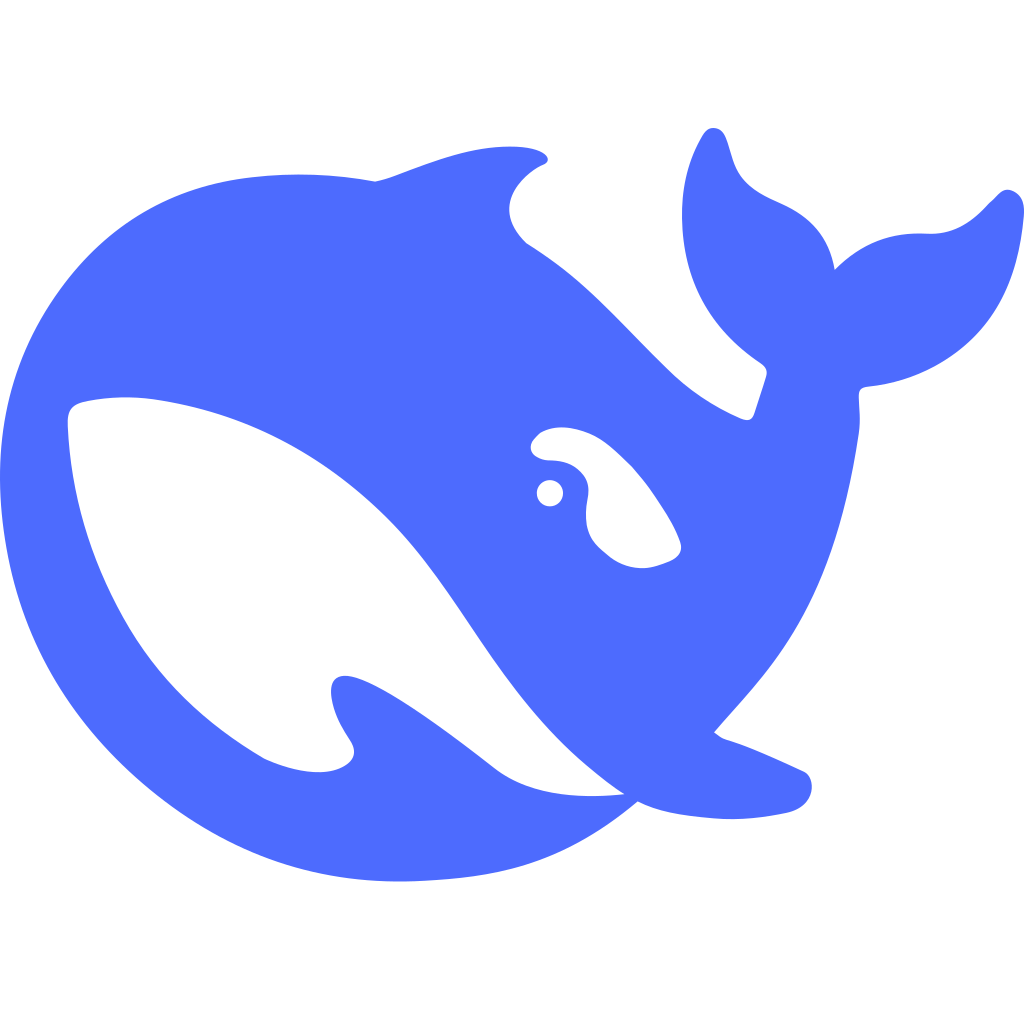}~~DeepSeek-70B & 65.20 & 75.20 & 81.03  & 86.67 & 64.92 & 72.60 \\
\includegraphics[height=1.2em]{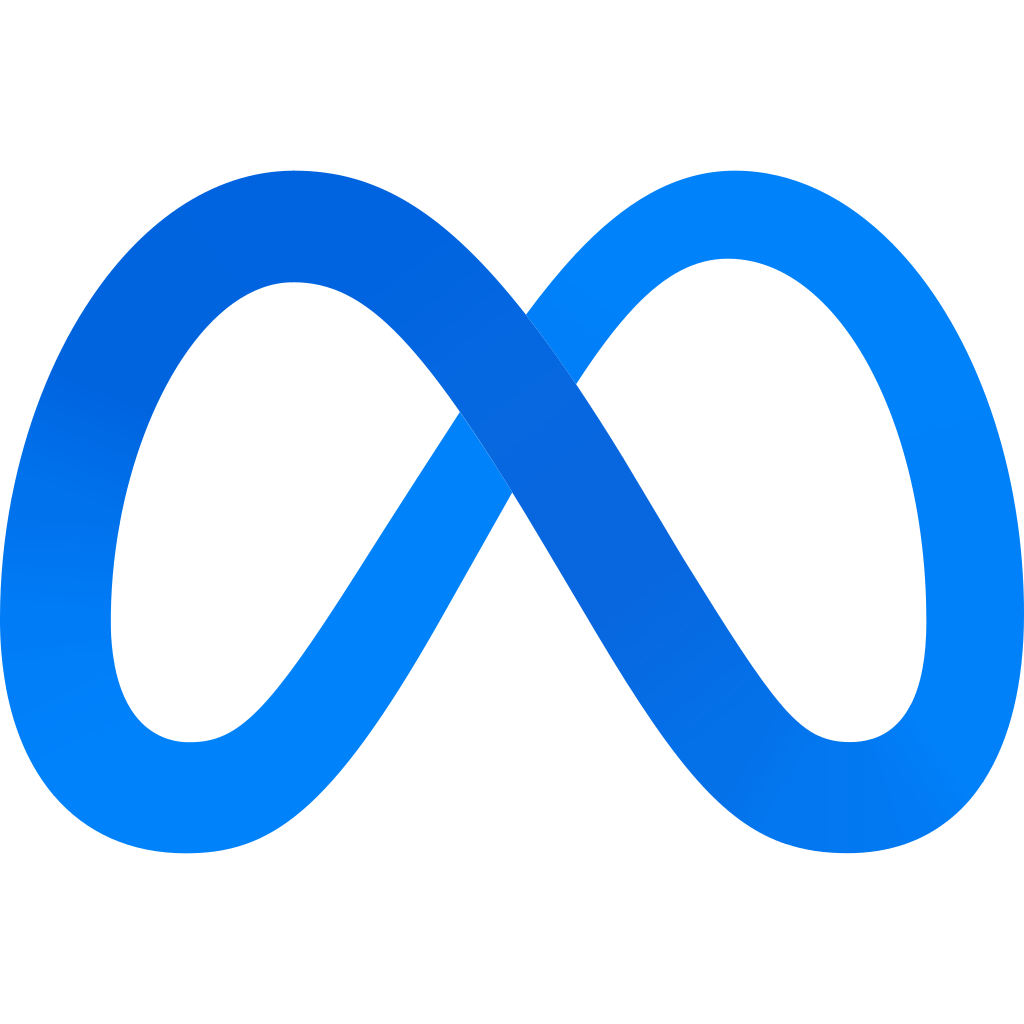}~~Meta-Llama3-8B & 66.78 & 75.09 & 82.51 & 87.63 & 67.39 & 74.59 \\
\includegraphics[height=.9em]{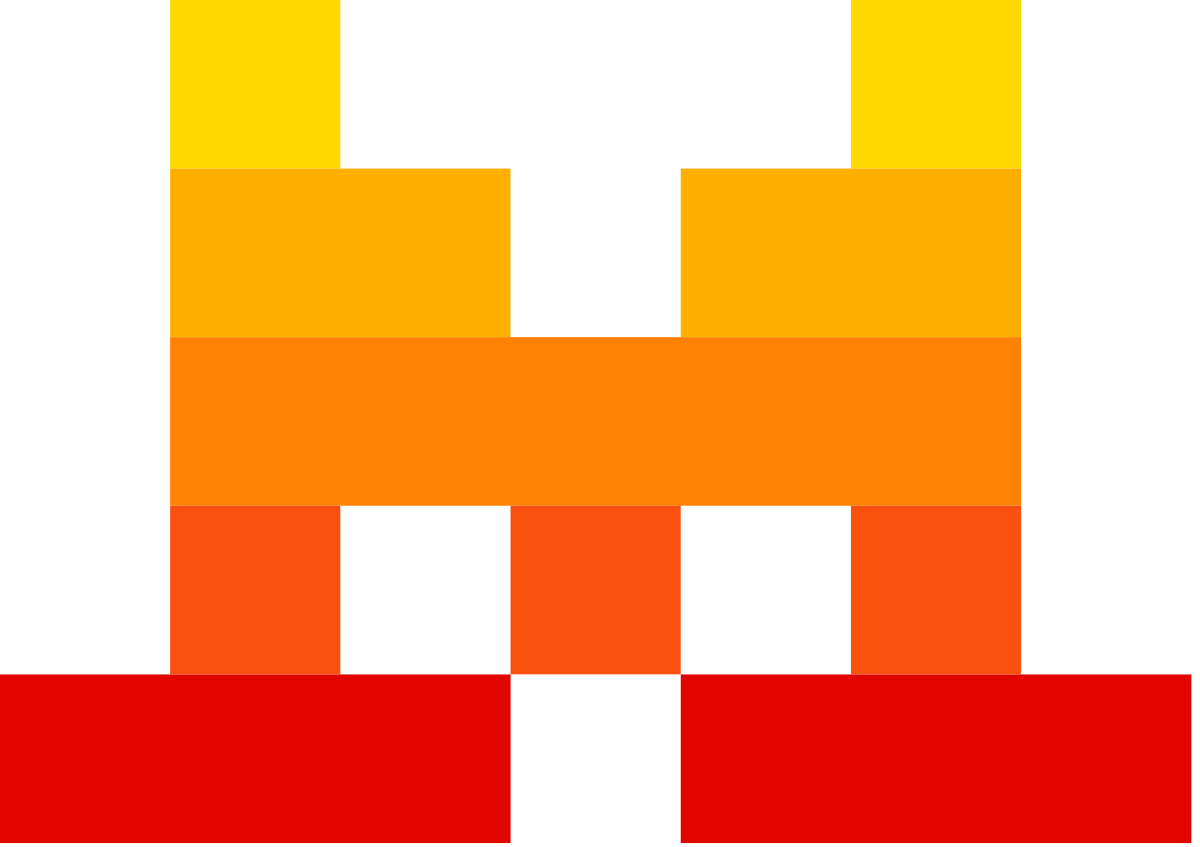}~~Mistral-7B   & \textbf{70.29} & 76.61 & 81.86 & 87.20 & 62.24 & 70.61 \\
\includegraphics[height=1.2em]{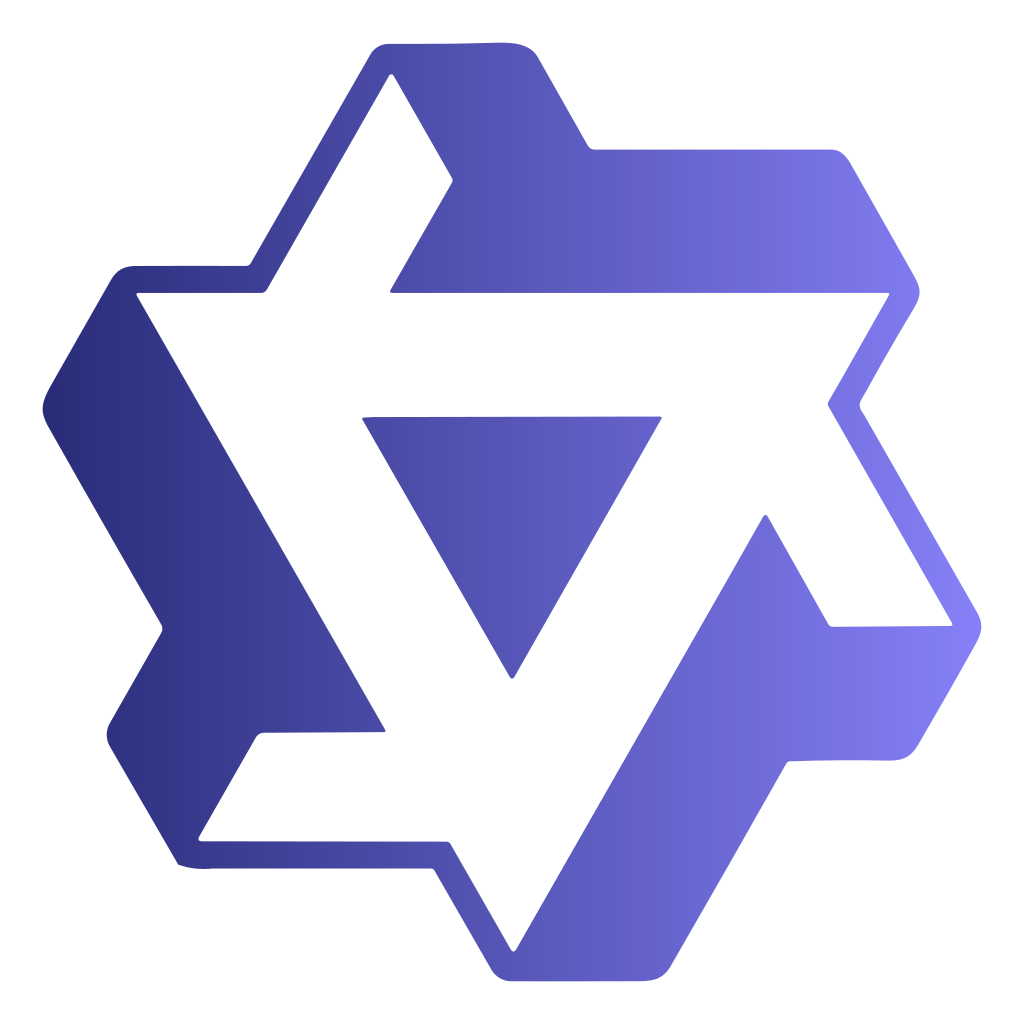}~~Qwen3-8B       & \underline{69.36} & 75.53 & 81.73 & 87.26 & 65.93 & 74.93 \\
\includegraphics[height=1.3em]{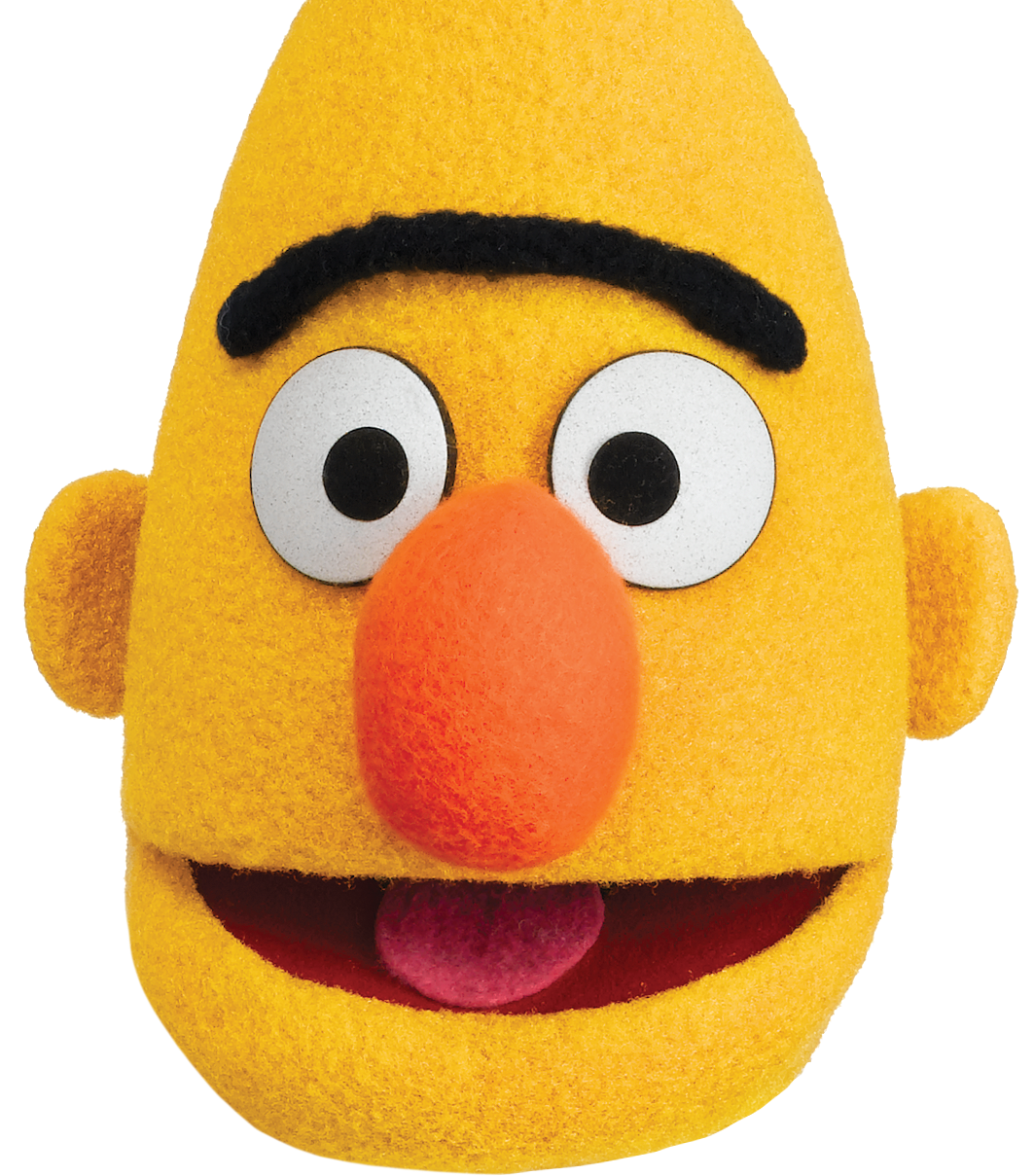}~~PCM (Ours)     & 65.75    & \underline{80.31}    & \textbf{88.86}    & \textbf{92.91}    & \textbf{71.10}   & \textbf{78.09}   \\
\includegraphics[height=1.3em]{images/Bert.png}~~PBR (Ours)     & 67.45    & \textbf{80.92}   & \underline{87.42}   & \underline{92.06}   & \underline{68.69}   & \underline{75.46}    \\
\bottomrule
\end{tabular}
}
\caption{Performance of LLMs Fine-Tuned with QLoRA vs. Prototype-Based Methods (Ours).}
\label{tab:llms}
\vspace{-1.3em}
\end{table}

%% file: tables/expert_evaluation.tex
\begin{table*}[ht]
\centering
\small
\arrayrulecolor{gray!30}
\renewcommand{\arraystretch}{1.15}
\resizebox{\linewidth}{!}{
\begin{tabular}{p{5cm} p{3.5cm} l l l c p{5cm}}
\toprule
\textbf{Input Excerpt} & \textbf{Confused Role Pair} & \textbf{Gold Label} & \textbf{Baseline Prediction} & \textbf{PCM Prediction} & \textbf{Error Reduction} & \textbf{Expert Assessment} \\
\midrule

\rowcolor{row1}
\textbf{As we have explained}, failure to comply with §262(l)(2)(A) is not an act of artificial infringement. 
& Recalling $\leftrightarrow$ Stating the Court’s reasoning 
& Recalling 
& Stating the Court’s reasoning 
& Recalling 
& 19.75\% 
& This case is ambiguous because “As we have explained” signals a backward reference, while the remainder conveys a definitive judicial conclusion, typical of Stating the Court’s reasoning. \\

\rowcolor{row2}
Thus, in Los Angeles Cloak Joint Board ILGWU (Helen Rose Co.), 127 N. L. R. B. 1543 (1960), \textbf{the Board held that} §8(b)(1)(B) barred a union from picketing a company in an attempt to force the employer to... 
& Describing $\leftrightarrow$ Recalling 
& Recalling 
& Describing 
& Recalling 
& 6.56\% 
& The difficulty arises because “the Board held that” points to recalling a precedent, yet the remainder paraphrases substantive content neutrally, resembling Describing. \\

\rowcolor{row3}
The American Bar Association \textbf{recommends} defense counsel “promptly communicate and explain to the defendant all plea offers made by the prosecuting attorney,” ABA Standards for Criminal Justice, Pleas of Guilty 14–3.2(a) (3d ed. 1999), 
& Quoting $\leftrightarrow$ Recalling 
& Recalling 
& Quoting 
& Recalling 
& 33.33\% 
& Ambiguity stems from combining an attribution verb (“recommends”) with a direct quotation, blending narrative recall with verbatim citation and blurring the distinction between Recalling and Quoting. \\

\bottomrule
\end{tabular}
}
\caption{Expert evaluation of model predictions on ambiguous rhetorical role pairs. PCM reduces baseline errors by better resolving overlaps between semantically similar functions.}
\label{tab:expert-eval}
\vspace{-1.3em}
\end{table*}

%% file: tables/hardness.tex
\begin{table}[t]
\centering
\small
\resizebox{\linewidth}{!}{%
\begin{tabular}{lcccc}
\toprule
\textbf{Model} &
\multicolumn{4}{c}{\textbf{Annotation Difficulty (Likert Scale)}} \\
\cmidrule(lr){2-5}
& \textbf{Easy} 
& \textbf{Rather easy} 
& \textbf{Rather difficult} 
& \textbf{Difficult} \\
\midrule
PCM\textsuperscript{\textcolor{baselinegold}{(Best)}}  
  & 100 & 79.62 & 61.53 & 16.66 \\
\bottomrule
\end{tabular}%
}
\caption{
Accuracy on hardness Likert scale annotated by experts.
}
\label{tab:difficulty-macro}
\vspace{-0.8em}
\end{table}

%% file: figures/meta-data-distr.tex
\begin{figure}[H]
    \centering
    \begin{subfigure}[t]{0.48\textwidth}
        \centering
        \includegraphics[width=\textwidth]{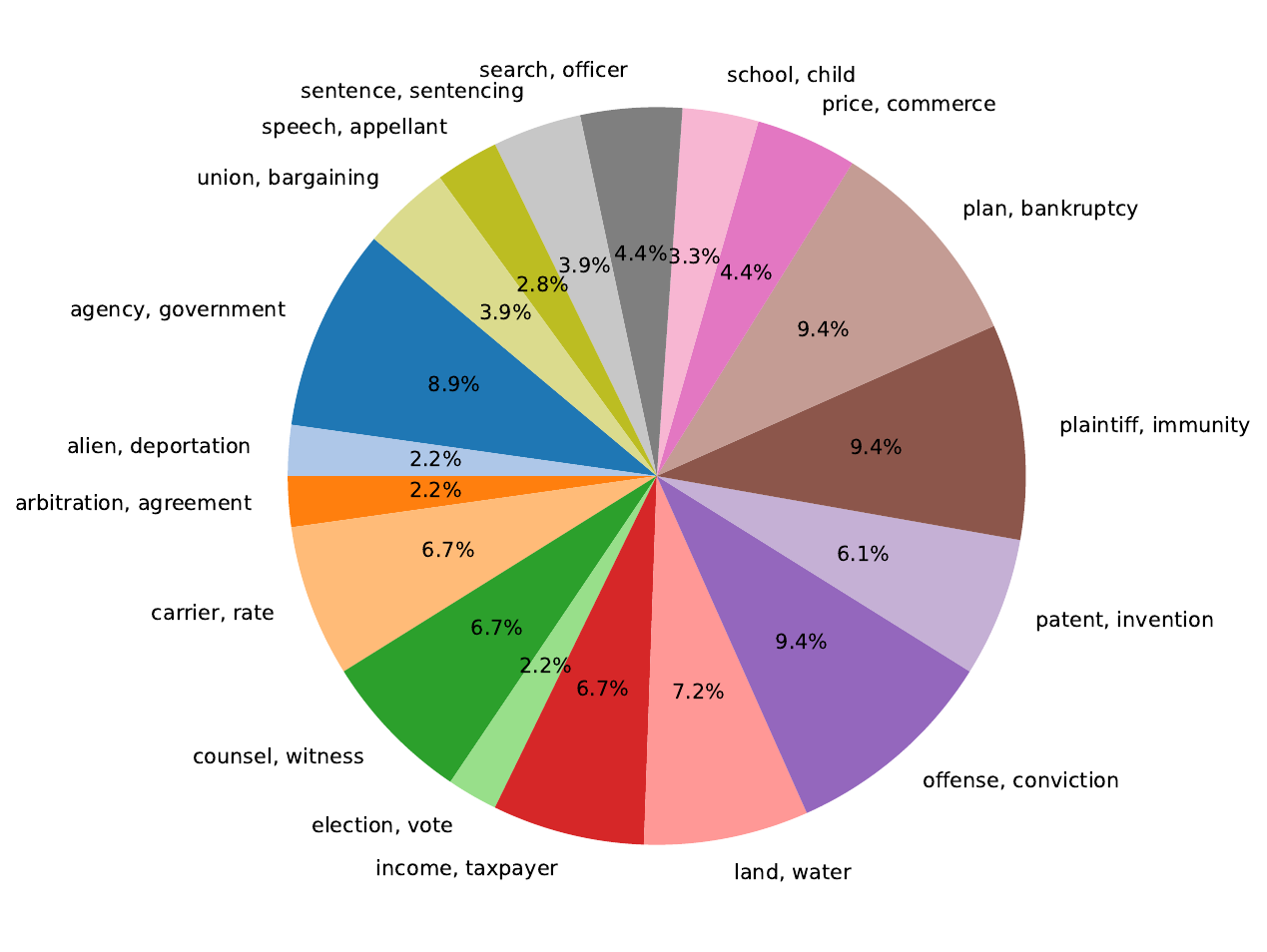}
    \end{subfigure}
    \hfill
    \begin{subfigure}[t]{0.48\textwidth}
        \centering
        \includegraphics[width=\textwidth]{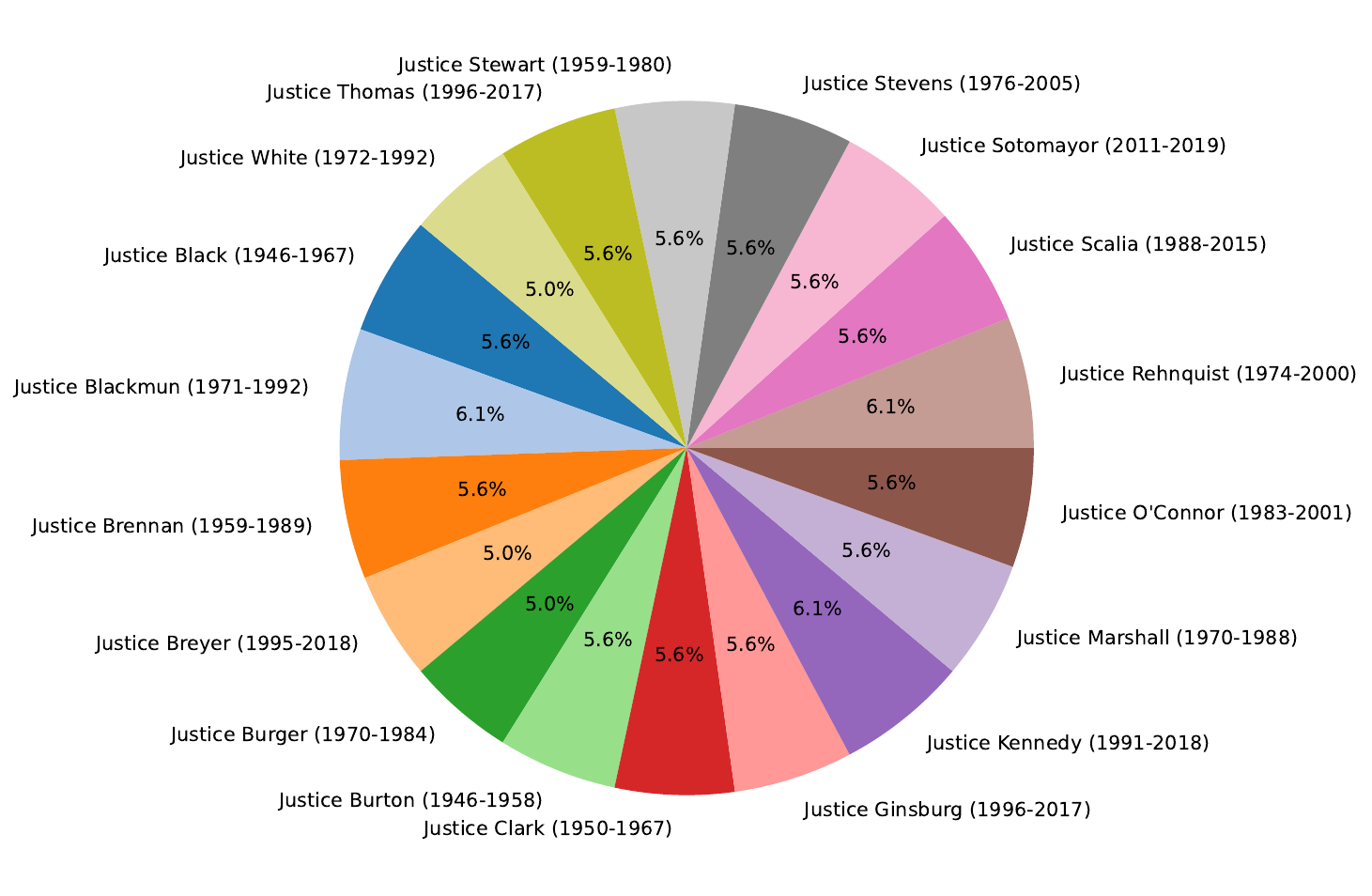}
    \end{subfigure}
    
    \caption{Topical, Temporal, and Authorial Diversity in our annotated corpus.}
    \label{fig:meta-stats}
\end{figure}

%% file: figures/position_rhetorical_functions.tex
\begin{figure}[h!]
\centering
\includegraphics[width=\linewidth]{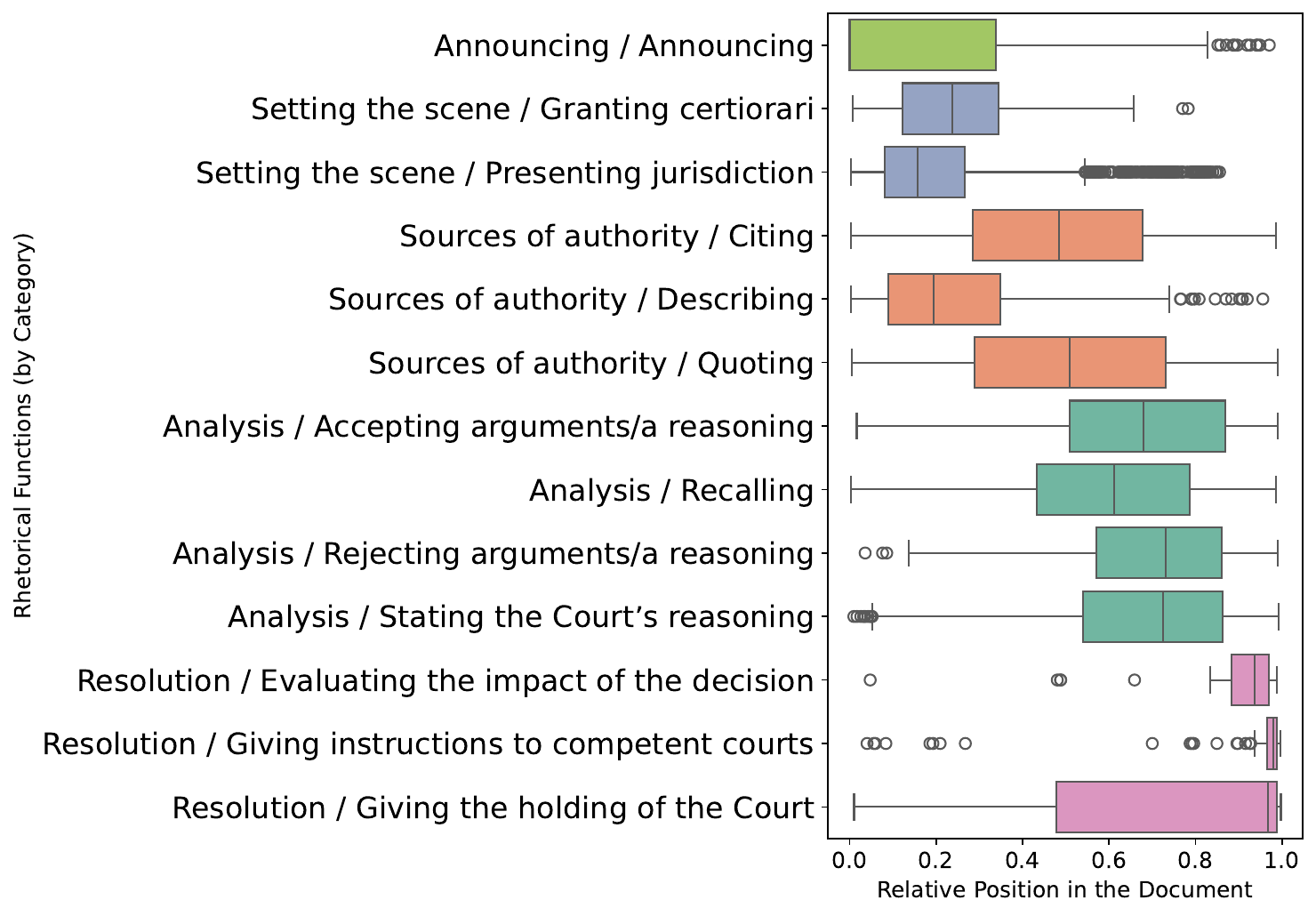}
\caption{Distribution of Rhetorical Functions by Relative Position, revealing a structured rhetorical flow in judicial reasoning—from the initial announcement to the final resolution.}
\label{fig:relative_pos}
\end{figure}

%% file: tables/stats-corpus.tex

\begin{table*}[!ht]\tiny
\centering
\begin{adjustbox}{max width=\textwidth}

\begin{tabular}{lrllrllrlrlrl}
\toprule
\textbf{Category}     & \textbf{} & \%(\(\downarrow\))
    & \textbf{Rhetorical Function}            & \textbf{} & \% (\(\downarrow\))
   & \textbf{Type}                                                                  & \textbf{} & \textbf{Target} & \textbf{} & \textbf{Author}       & \textbf{} & \%(\(\rightarrow\))
    \\ \midrule
Announcing           & 344       & 1.30  & Announcing                              & 344       & 1.30  &                                                                                &           &                 &           &                       &           & 1.30  \\ \midrule
Setting the scene    & 5.123     & 19.45 & Granting certiorari                     & 182       & 0.69  &                                                                                &           &                 &           &                       &           & 0.69  \\
                     &           &       & Presenting jurisdiction                 & 4.941     & 18.76 & Adjudicated facts                                                              & 2.283     &                 &           &                       &           & 8.67  \\
                     &           &       &                                         &           &       & Lower court decision                                                           & 1.192     &                 &           &                       &           & 4.52  \\
                     &           &       &                                         &           &       & Context                                                                        & 467       &                 &           &                       &           & 1.77  \\
                     &           &       &                                         &           &       & Other procedural events                                                        & 412       &                 &           &                       &           & 1.56  \\
                     &           &       &                                         &           &       & \begin{tabular}[c]{@{}l@{}}Parties' legal claims \\ and arguments\end{tabular} & 363       &                 &           &                       &           & 1.37  \\
                     &           &       &                                         &           &       & Legal question(s)                                                              & 224       &                 &           &                       &           & 0.85  \\ \midrule
Sources of authority & 8.041     & 30.54 & Citing                                  & 6.442     & 2.44  & SCOTUS decision                                                                & 2.764     &                 &           &                       &           & 0.89  \\
                     &           &       &                                         &           &       & Primary source of law                                                          & 2.203     &                 &           &                       &           & 0.91  \\
                     &           &       &                                         &           &       & Secondary source of law                                                        & 1.474     &                 &           &                       &           & 0.63  \\
                     &           &       & Describing                              & 955       & 3.62  & Primary source of law                                                          & 771       &                 &           &                       &           & 2.92  \\
                     &           &       &                                         &           &       & Secondary source of law                                                        & 159       &                 &           &                       &           & 0.60  \\
                     &           &       &                                         &           &       & \multirow{2}{2cm}{Established practices  or cultural norms}                      & 25        &                 &           &                       &           & 0.09  \\
                     &           &       &                                         &           &       &                                                                                &           &                 &           &                       &           &       \\
                     &           &       & Quoting                                 & 644       & 24.46 & SCOTUS decision                                                                & 235       &                 &           &                       &           & 10.49 \\
                     &           &       &                                         &           &       & Primary source of law                                                          & 241       &                 &           &                       &           & 8.36  \\
                     &           &       &                                         &           &       & Secondary source of law                                                        & 168       &                 &           &                       &           & 5.59  \\ \midrule
Analysis             & 11.910    & 45.23 & Stating the Court's reasoning           & 3.198     & 12.14 &                                                                                &           &                 &           &                       &           & 12.14 \\
                     &           &       & Rejecting arguments/a reasoning         & 490       & 1.86  &                                                                                &           &                 &           &                       &           & 1.86  \\
                     &           &       & Accepting arguments/a reasoning         & 103       & 0.39  &                                                                                &           &                 &           &                       &           & 0.39  \\
                     &           &       & Recalling                               & 8.119     & 30.83 & A SCOTUS opinion                                                               & 2.160     &                 &           &                       &           & 8.20  \\
                     &           &       &                                         &           &       & A primary source                                                               & 1.781     &                 &           &                       &           & 6.76  \\
                     &           &       &                                         &           &       & A secondary source                                                             & 359       &                 &           &                       &           & 1.36  \\
                     &           &       &                                         &           &       & \multirow{2}{2cm}{An established practice  or cultural norm}                      & 1.199     &                 &           &                       &           & 4.55  \\
                     &           &       &                                         &           &       &                                                                                &           &                 &           &                       &           &       \\
                     &           &       &                                         &           &       & \multirow{2}{2cm}{An adjudicated fact or procedural event}                       & 1.447     & Present case    & 1.152     &                       &           & 4.37  \\
                     &           &       &                                         &           &       &                                                                                &           &                 &           &                       &           &       \\
                     &           &       &                                         &           &       &                                                                                &           & Another case    & 295       &                       &           & 1.12  \\
                     &           &       &                                         &           &       & Legal question(s)                                                              & 182       & Present case    & 147       &                       &           & 0.55  \\
                     &           &       &                                         &           &       &                                                                                &           & Another case    & 35        &                       &           & 0.13  \\
                     &           &       &                                         &           &       & An argument                                                                    & 991       & Present case    & 967       & Petitioner            & 413       & 1.64  \\
                     &           &       &                                         &           &       &                                                                                &           &                 &           & Respondent            & 513       & 1.94  \\
                     &           &       &                                         &           &       &                                                                                &           &                 &           & Dissenting justice(s) & 22        & 0.08  \\
                     &           &       &                                         &           &       &                                                                                &           & Another case    & 24        &                       &           & 0.09  \\ \midrule
Resolution           & 910       & 3.45  & Giving the holding of the Court         & 760       & 2.88  &                                                                                &           &                 &           &                       &           & 2.88  \\
                     &           &       & Giving instructions to competent courts & 105       & 0.39  &                                                                                &           &                 &           &                       &           & 0.39  \\
                     &           &       & Evaluating the impact of the decision   & 45        & 0.17  &                                                                                &           &                 &           &                       &           & 0.17  \\ \midrule
\textbf{Total}       & 26.328    &       &                                         &           &       &                                                                                &           &                 &           &                       &           &       \\ \midrule
\end{tabular}
\end{adjustbox}
\caption{Final Annotation Scheme: Comprising 5 Categories, 13 Rhetorical Functions, and 24 Attributes (Types, Targets, and Authors). Counts of Text Segments are Provided, with Distributions Displayed at the Category Level (\(\downarrow\)), Rhetorical Function Level (\(\downarrow\)), and Step Level (\(\rightarrow\)).}
\label{tab:codingscheme_and_number_of_annotations}
\end{table*}

%% file: figures/tagset.tex
\begin{figure*}[!ht]
    \centering
    \includegraphics[width=0.7\textwidth]{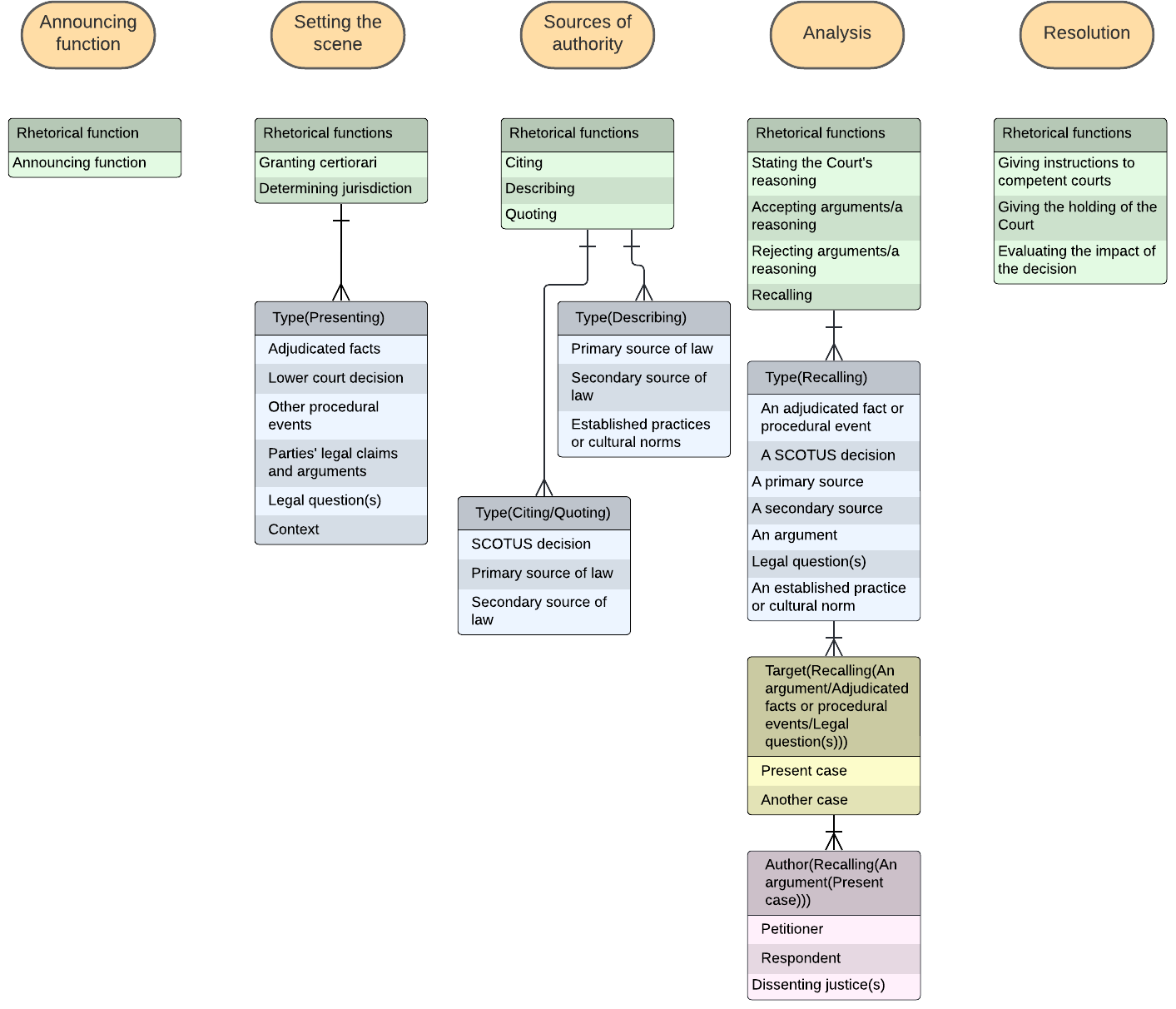}
    \caption{The final coding scheme is composed of 5 categories (ovals with orange background), 13 rhetorical functions (green rectangles) and 24 attributes (types in blue rectangles, target in the yellow rectangle, and author in the purple rectangle. The scheme reads from top to bottom: A step label is constructed by first choosing a category, then a rhetorical function, then if required, by combining attributes to complete the discursive information provided by the rhetorical function.}

        \label{fig:final_scheme}
\end{figure*}

%% file: figures/hierarchical_architecture.tex
\begin{figure}[h!]
\centering
\includegraphics[width=\linewidth]{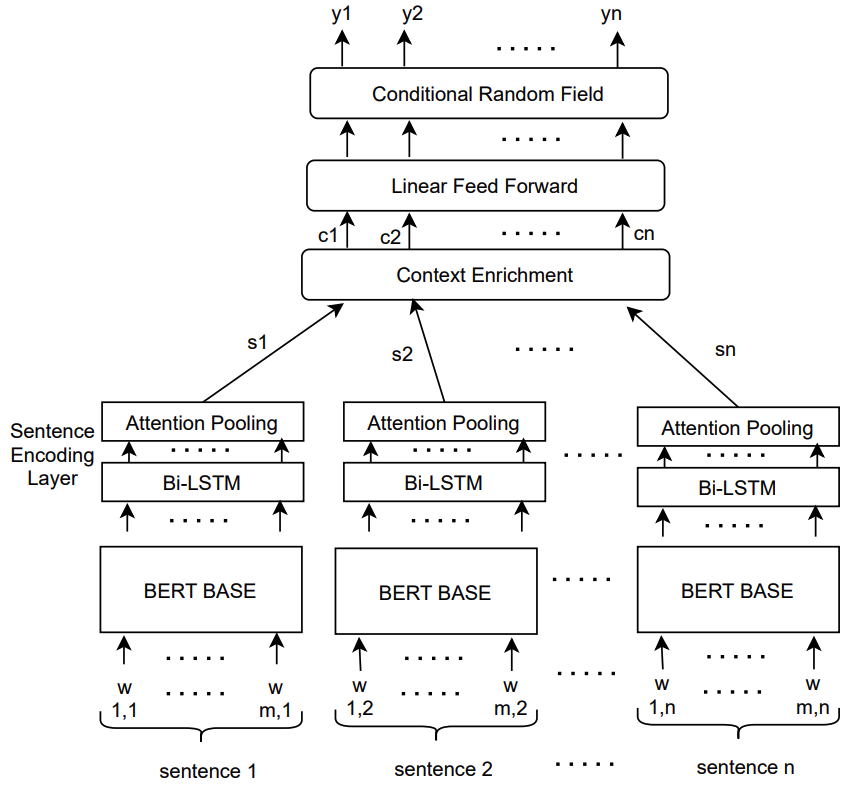}
\caption{The hierarchical architecture.}
\label{fig:hierarchical-architecture}
\end{figure}

%% file: tables/datasets_details.tex
\begin{table*}[t]
\centering
\small
\arrayrulecolor{gray!30}
\rowcolors{2}{white}{gray!08}
\resizebox{\linewidth}{!}{%
\begin{tabular}{lcccccc}
\toprule
\textbf{Dataset} & \textbf{Source} & \textbf{Domain} & \textbf{Language} & \textbf{\# Docs} & \textbf{\# Sents} & \textbf{Labels} \\
\midrule
\textsc{Scotus}\textsubscript{Category}         & Ours & Legal (U.S.)     & English & 180 & 26,327 & 5 \\
\textsc{Scotus}\textsubscript{RF}    & Ours & Legal (U.S.)     & English & 180 & 26,327 & 13 \\
\textsc{Scotus}\textsubscript{Steps}          & Ours & Legal (U.S.)     & English & 180 & 26,327 & 35 \\
\textsc{LegalEval}                    & \citet{kalamkar-etal-2022-corpus} & Legal (India)   & English & 214 & 31,865 & 13 \\
\textsc{DeepRhole}                     & \citet{bhattacharya2023deeprhole} & Legal (India)   & English & 50  & 9,380  & 7 \\
PubMed                         & \citet{dernoncourt-lee-2017-pubmed} & Medical         & English & 20,000 & 227,000 & 5 \\
\textsc{CS-Abstracts}                   & \citet{gonçalves_2020} & Scientific & English & 654 & 7,385 & 5 \\
\bottomrule
\end{tabular}
}
\caption{Evaluation datasets used in our experiments. \textsc{Scotus} is annotated at three hierarchical levels: category, rhetorical function, and steps.}
\label{tab:evaluation-datasets}
\end{table*}

%% file: tables/injection.tex
\begin{table}[t!]
\centering
\small
\arrayrulecolor{gray!30}
\rowcolors{2}{white}{gray!08}
\resizebox{\linewidth}{!}{%
\begin{tabular}{>{\bfseries}lccc}
\toprule
Method & \textsc{Scotus}\textsubscript{RF} & \textsc{LegalEval} & \textsc{PubMed} \\
\midrule
Linear Fusion                & 80.89  &  91.62 & 91.91 \\
Conditional Layer Norm & 78.11 & 87.49 & 92.74 \\
Cross-Attention Fusion       & 79.30 & 87.74 & 92.20 \\
Feature-wise Linear Mod. & 74.71 & 76.74 & 92.74 \\
Gated Residual Addition      & 79.58 & 89.06 & 92.79 \\
\bottomrule
\end{tabular}%
}
\caption{W-F1 scores for prototype injection strategies. All variants share the same hierarchical encoder with PCM integration.}
\label{tab:pcm-injection-wf1}
  \vspace{-1.4em}

\end{table}

%% file: figures/protototype_sensitivity.tex
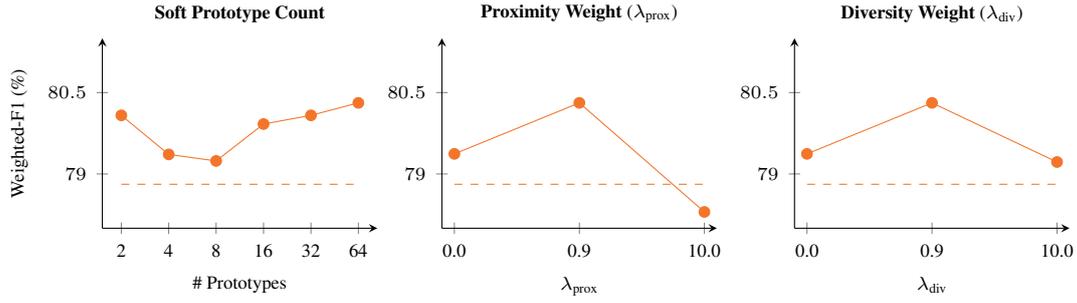
\begin{figure*}[t!]
\centering
\begin{tikzpicture}

\begin{axis}[
    name=rf_proto,
    width=5.2cm, height=4.1cm,
    ymin=78.0, ymax=81.5,
    ytick={79.0, 80.5},
    xtick=data,
    xticklabels={2, 4, 8, 16, 32, 64},
    xmin=-0.4, xmax=5.4,
    xlabel={\scriptsize \# Prototypes},
    ylabel={\scriptsize Weighted-F1 (\%)},
    tick label style={font=\scriptsize},
    label style={font=\scriptsize},
    title={\scriptsize \textbf{Soft Prototype Count}},
    title style={yshift=-0.8ex},
    axis lines=left,
    clip=false,
]
\addplot[myproto, mark=*] coordinates {(0,80.08) (1,79.36) (2,79.24) (3,79.92) (4,80.08) (5,80.31)};
\addplot[myproto, dashed] coordinates {(0,78.81) (5,78.81)};
\end{axis}

\begin{axis}[
    at={(rf_proto.right of south east)}, anchor=left of south west,
    name=rf_lambdap,
    width=5.2cm, height=4.1cm,
    ymin=78.0, ymax=81.5,
    ytick={79.0, 80.5},
    xtick=data,
    xticklabels={0.0, 0.9, 10.0},
    xlabel={\scriptsize $\lambda_{\text{prox}}$},
    xmin=-0.1, xmax=2.1,
    tick label style={font=\scriptsize},
    label style={font=\scriptsize},
    title={\scriptsize \textbf{Proximity Weight} (\(\lambda_{\text{prox}}\))},
    title style={yshift=-0.8ex},
    axis lines=left,
    clip=false,
]
\addplot[myproto, mark=*] coordinates {(0,79.37) (1,80.31) (2,78.3)};
\addplot[myproto, dashed] coordinates {(0,78.81) (2,78.81)};
\end{axis}

\begin{axis}[
    at={(rf_lambdap.right of south east)}, anchor=left of south west,
    width=5.2cm, height=4.1cm,
    ymin=78.0, ymax=81.5,
    ytick={79.0, 80.5},
    xtick=data,
    xticklabels={0.0, 0.9, 10.0},
    xlabel={\scriptsize $\lambda_{\text{div}}$},
    xmin=-0.1, xmax=2.1,
    tick label style={font=\scriptsize},
    label style={font=\scriptsize},
    title={\scriptsize \textbf{Diversity Weight} (\(\lambda_{\text{div}}\))},
    title style={yshift=-0.8ex},
    axis lines=left,
    clip=false,
]
\addplot[myproto, mark=*] coordinates {(0,79.37) (1,80.31) (2,79.22)};
\addplot[myproto, dashed] coordinates {(0,78.81) (2,78.81)};
\end{axis}

\end{tikzpicture}
\vspace{-0.6em}
\caption{
\small
Effect of PBR hyperparameters on w-F1 at the  \textsc{SCOTUS}\textsubscript{RF} 
Dashed lines indicate the baseline without prototypes.  
}
\label{fig:rf-proto-lambdas-final}
\vspace{-0.9em}

\end{figure*}

%% file: custom.bib
@inproceedings{neves-etal-2019-evaluation,
    title = "Evaluation of Scientific Elements for Text Similarity in Biomedical Publications",
    author = "Neves, Mariana  and
      Butzke, Daniel  and
      Grune, Barbara",
    editor = "Stein, Benno  and
      Wachsmuth, Henning",
    booktitle = "Proceedings of the 6th Workshop on Argument Mining",
    month = aug,
    year = "2019",
    address = "Florence, Italy",
    publisher = "Association for Computational Linguistics",
    url = "https://aclanthology.org/W19-4515/",
    doi = "10.18653/v1/W19-4515",
    pages = "124--135"
}

@article{safder2019bibliometric,
  title={Bibliometric-enhanced information retrieval: a novel deep feature engineering approach for algorithm searching from full-text publications},
  author={Safder, Iqra and Hassan, Saeed-Ul},
  journal={Scientometrics},
  volume={119},
  pages={257--277},
  year={2019},
  publisher={Springer}
}

@inproceedings{kalamkar-etal-2022-corpus,
    title = "Corpus for Automatic Structuring of Legal Documents",
    author = "Kalamkar, Prathamesh  and
      Tiwari, Aman  and
      Agarwal, Astha  and
      Karn, Saurabh  and
      Gupta, Smita  and
      Raghavan, Vivek  and
      Modi, Ashutosh",
    editor = "Calzolari, Nicoletta  and
      B{\'e}chet, Fr{\'e}d{\'e}ric  and
      Blache, Philippe  and
      Choukri, Khalid  and
      Cieri, Christopher  and
      Declerck, Thierry  and
      Goggi, Sara  and
      Isahara, Hitoshi  and
      Maegaard, Bente  and
      Mariani, Joseph  and
      Mazo, H{\'e}l{\`e}ne  and
      Odijk, Jan  and
      Piperidis, Stelios",
    booktitle = "Proceedings of the Thirteenth Language Resources and Evaluation Conference",
    month = jun,
    year = "2022",
    address = "Marseille, France",
    publisher = "European Language Resources Association",
    url = "https://aclanthology.org/2022.lrec-1.470/",
    pages = "4420--4429"
}

@inproceedings{muhammed2024impact,
  title={Impact of Rhetorical Roles in Abstractive Legal Document Summarization},
  author={Muhammed, Akheel and Muslihuddeen, Hamna and Sankar, Shalaka and Kumar, M Anand},
  booktitle={2024 5th International Conference on Innovative Trends in Information Technology (ICITIIT)},
  pages={1--6},
  year={2024},
  organization={IEEE}
}

@article{walker2019automatic,
  title={Automatic Classification of Rhetorical Roles for Sentences: Comparing Rule-Based Scripts with Machine Learning.},
  author={Walker, Vern R and Pillaipakkamnatt, Krishnan and Davidson, Alexandra M and Linares, Marysa and Pesce, Domenick J},
  journal={ASAIL@ ICAIL},
  volume={2385},
  year={2019}
}

@inproceedings{brack2022cross,
  title={Cross-domain multi-task learning for sequential sentence classification in research papers},
  author={Brack, Arthur and Hoppe, Anett and Buscherm{\"o}hle, Pascal and Ewerth, Ralph},
  booktitle={Proceedings of the 22nd ACM/IEEE Joint Conference on Digital Libraries},
  pages={1--13},
  year={2022}
}

@article{bhattacharya_deeprhole_2023,
	title = {{DeepRhole}: deep learning for rhetorical role labeling of sentences in legal case documents},
	volume = {31},
	issn = {1572-8382},
	url = {https://doi.org/10.1007/s10506-021-09304-5},
	doi = {10.1007/s10506-021-09304-5},
	number = {1},
	journal = {Artificial Intelligence and Law},
	author = {Bhattacharya, Paheli and Paul, Shounak and Ghosh, Kripabandhu and Ghosh, Saptarshi and Wyner, Adam},
	month = mar,
	year = {2023},
	pages = {53--90},
}

@inproceedings{nigam-etal-2025-legalseg,
    title = "{L}egal{S}eg: Unlocking the Structure of {I}ndian Legal Judgments Through Rhetorical Role Classification",
    author = "Nigam, Shubham Kumar  and
      Dubey, Tanmay  and
      Sharma, Govind  and
      Shallum, Noel  and
      Ghosh, Kripabandhu  and
      Bhattacharya, Arnab",
    editor = "Chiruzzo, Luis  and
      Ritter, Alan  and
      Wang, Lu",
    booktitle = "Findings of the Association for Computational Linguistics: NAACL 2025",
    month = apr,
    year = "2025",
    address = "Albuquerque, New Mexico",
    publisher = "Association for Computational Linguistics",
    url = "https://aclanthology.org/2025.findings-naacl.63/",
    pages = "1129--1144",
    ISBN = "979-8-89176-195-7"
}

@inproceedings{snell2017,
 author = {Snell, Jake and Swersky, Kevin and Zemel, Richard},
 booktitle = {Advances in Neural Information Processing Systems},
 editor = {I. Guyon and U. Von Luxburg and S. Bengio and H. Wallach and R. Fergus and S. Vishwanathan and R. Garnett},
 pages = {},
 publisher = {Curran Associates, Inc.},
 title = {Prototypical Networks for Few-shot Learning},
 url = {https://proceedings.neurips.cc/paper_files/paper/2017/file/cb8da6767461f2812ae4290eac7cbc42-Paper.pdf},
 volume = {30},
 year = {2017}
}

@inproceedings{huang-etal-2023-pram,
    title = "{PRAM}: An End-to-end Prototype-based Representation Alignment Model for Zero-resource Cross-lingual Named Entity Recognition",
    author = "Huang, Yucheng  and
      Liu, Wenqiang  and
      Zhang, Xianli  and
      Lang, Jun  and
      Gong, Tieliang  and
      Li, Chen",
    editor = "Rogers, Anna  and
      Boyd-Graber, Jordan  and
      Okazaki, Naoaki",
    booktitle = "Findings of the Association for Computational Linguistics: ACL 2023",
    month = jul,
    year = "2023",
    address = "Toronto, Canada",
    publisher = "Association for Computational Linguistics",
    url = "https://aclanthology.org/2023.findings-acl.201/",
    doi = "10.18653/v1/2023.findings-acl.201",
    pages = "3220--3233"
}

@inproceedings{yu-etal-2022-dependency,
    title = "Dependency-aware Prototype Learning for Few-shot Relation Classification",
    author = "Yu, Tianshu  and
      Yang, Min  and
      Zhao, Xiaoyan",
    editor = "Calzolari, Nicoletta  and
      Huang, Chu-Ren  and
      Kim, Hansaem  and
      Pustejovsky, James  and
      Wanner, Leo  and
      Choi, Key-Sun  and
      Ryu, Pum-Mo  and
      Chen, Hsin-Hsi  and
      Donatelli, Lucia  and
      Ji, Heng  and
      Kurohashi, Sadao  and
      Paggio, Patrizia  and
      Xue, Nianwen  and
      Kim, Seokhwan  and
      Hahm, Younggyun  and
      He, Zhong  and
      Lee, Tony Kyungil  and
      Santus, Enrico  and
      Bond, Francis  and
      Na, Seung-Hoon",
    booktitle = "Proceedings of the 29th International Conference on Computational Linguistics",
    month = oct,
    year = "2022",
    address = "Gyeongju, Republic of Korea",
    publisher = "International Committee on Computational Linguistics",
    url = "https://aclanthology.org/2022.coling-1.205/",
    pages = "2339--2345"
}

@inproceedings{luo-etal-2023-prototype,
    title = "Prototype-Based Interpretability for Legal Citation Prediction",
    author = "Luo, Chu Fei  and
      Bhambhoria, Rohan  and
      Dahan, Samuel  and
      Zhu, Xiaodan",
    editor = "Rogers, Anna  and
      Boyd-Graber, Jordan  and
      Okazaki, Naoaki",
    booktitle = "Findings of the Association for Computational Linguistics: ACL 2023",
    month = jul,
    year = "2023",
    address = "Toronto, Canada",
    publisher = "Association for Computational Linguistics",
    url = "https://aclanthology.org/2023.findings-acl.301/",
    doi = "10.18653/v1/2023.findings-acl.301",
    pages = "4883--4898"
}

@article{ruch2007using,
  title={Using argumentation to extract key sentences from biomedical abstracts},
  author={Ruch, Patrick and Boyer, Celia and Chichester, Christine and Tbahriti, Imad and Geissb{\"u}hler, Antoine and Fabry, Paul and Gobeill, Julien and Pillet, Violaine and Rebholz-Schuhmann, Dietrich and Lovis, Christian and others},
  journal={International journal of medical informatics},
  volume={76},
  number={2-3},
  pages={195--200},
  year={2007},
  publisher={Elsevier}
}

@inproceedings{mcknight2003categorization,
  title={Categorization of sentence types in medical abstracts},
  author={McKnight, Larry and Srinivasan, Padmini},
  booktitle={AMIA annual symposium proceedings},
  volume={2003},
  pages={440},
  year={2003}
}

@inproceedings{lin-etal-2006-generative,
    title = "Generative Content Models for Structural Analysis of Medical Abstracts",
    author = "Lin, Jimmy  and
      Karakos, Damianos  and
      Demner-Fushman, Dina  and
      Khudanpur, Sanjeev",
    editor = "Verspoor, Karin  and
      Cohen, Kevin Bretonnel  and
      Goertzel, Ben  and
      Mani, Inderjeet",
    booktitle = "Proceedings of the {HLT}-{NAACL} {B}io{NLP} Workshop on Linking Natural Language and Biology",
    month = jun,
    year = "2006",
    address = "New York, New York",
    publisher = "Association for Computational Linguistics",
    url = "https://aclanthology.org/W06-3309/",
    pages = "65--72"
}

@inproceedings{cohan-etal-2019-pretrained,
    title = "Pretrained Language Models for Sequential Sentence Classification",
    author = "Cohan, Arman  and
      Beltagy, Iz  and
      King, Daniel  and
      Dalvi, Bhavana  and
      Weld, Dan",
    editor = "Inui, Kentaro  and
      Jiang, Jing  and
      Ng, Vincent  and
      Wan, Xiaojun",
    booktitle = "Proceedings of the 2019 Conference on Empirical Methods in Natural Language Processing and the 9th International Joint Conference on Natural Language Processing (EMNLP-IJCNLP)",
    month = nov,
    year = "2019",
    address = "Hong Kong, China",
    publisher = "Association for Computational Linguistics",
    url = "https://aclanthology.org/D19-1383/",
    doi = "10.18653/v1/D19-1383",
    pages = "3693--3699"
}

@inproceedings{devlin-etal-2019-bert,
    title = "{BERT}: Pre-training of Deep Bidirectional Transformers for Language Understanding",
    author = "Devlin, Jacob  and
      Chang, Ming-Wei  and
      Lee, Kenton  and
      Toutanova, Kristina",
    editor = "Burstein, Jill  and
      Doran, Christy  and
      Solorio, Thamar",
    booktitle = "Proceedings of the 2019 Conference of the North {A}merican Chapter of the Association for Computational Linguistics: Human Language Technologies, Volume 1 (Long and Short Papers)",
    month = jun,
    year = "2019",
    address = "Minneapolis, Minnesota",
    publisher = "Association for Computational Linguistics",
    url = "https://aclanthology.org/N19-1423/",
    doi = "10.18653/v1/N19-1423",
    pages = "4171--4186"
}

@inproceedings{jin-szolovits-2018-hierarchical,
    title = "Hierarchical Neural Networks for Sequential Sentence Classification in Medical Scientific Abstracts",
    author = "Jin, Di  and
      Szolovits, Peter",
    editor = "Riloff, Ellen  and
      Chiang, David  and
      Hockenmaier, Julia  and
      Tsujii, Jun{'}ichi",
    booktitle = "Proceedings of the 2018 Conference on Empirical Methods in Natural Language Processing",
    month = oct # "-" # nov,
    year = "2018",
    address = "Brussels, Belgium",
    publisher = "Association for Computational Linguistics",
    url = "https://aclanthology.org/D18-1349/",
    doi = "10.18653/v1/D18-1349",
    pages = "3100--3109"
}

@article{brack2024sequential,
  title={Sequential sentence classification in research papers using cross-domain multi-task learning},
  author={Brack, Arthur and Entrup, Elias and Stamatakis, Markos and Buscherm{\"o}hle, Pascal and Hoppe, Anett and Ewerth, Ralph},
  journal={International Journal on Digital Libraries},
  volume={25},
  number={2},
  pages={377--400},
  year={2024},
  publisher={Springer}
}

@inproceedings{teufel-etal-2009-towards,
    title = "Towards Domain-Independent Argumentative Zoning: Evidence from Chemistry and Computational Linguistics",
    author = "Teufel, Simone  and
      Siddharthan, Advaith  and
      Batchelor, Colin",
    editor = "Koehn, Philipp  and
      Mihalcea, Rada",
    booktitle = "Proceedings of the 2009 Conference on Empirical Methods in Natural Language Processing",
    month = aug,
    year = "2009",
    address = "Singapore",
    publisher = "Association for Computational Linguistics",
    url = "https://aclanthology.org/D09-1155/",
    pages = "1493--1502"
}

@inproceedings{feng-hirst-2012-text,
    title = "Text-level Discourse Parsing with Rich Linguistic Features",
    author = "Feng, Vanessa Wei  and
      Hirst, Graeme",
    editor = "Li, Haizhou  and
      Lin, Chin-Yew  and
      Osborne, Miles  and
      Lee, Gary Geunbae  and
      Park, Jong C.",
    booktitle = "Proceedings of the 50th Annual Meeting of the Association for Computational Linguistics (Volume 1: Long Papers)",
    month = jul,
    year = "2012",
    address = "Jeju Island, Korea",
    publisher = "Association for Computational Linguistics",
    url = "https://aclanthology.org/P12-1007/",
    pages = "60--68"
}

@inproceedings{t-y-s-s-etal-2024-mind,
    title = "Mind Your Neighbours: Leveraging Analogous Instances for Rhetorical Role Labeling for Legal Documents",
    author = "T.y.s.s., Santosh  and
      Sarwat, Hassan  and
      Abdou, Ahmed Mohamed Abdelaal  and
      Grabmair, Matthias",
    editor = "Calzolari, Nicoletta  and
      Kan, Min-Yen  and
      Hoste, Veronique  and
      Lenci, Alessandro  and
      Sakti, Sakriani  and
      Xue, Nianwen",
    booktitle = "Proceedings of the 2024 Joint International Conference on Computational Linguistics, Language Resources and Evaluation (LREC-COLING 2024)",
    month = may,
    year = "2024",
    address = "Torino, Italia",
    publisher = "ELRA and ICCL",
    url = "https://aclanthology.org/2024.lrec-main.987/",
    pages = "11296--11306"
}

@inproceedings{t-y-s-s-etal-2024-hiculr,
    title = "{H}i{C}u{LR}: Hierarchical Curriculum Learning for Rhetorical Role Labeling of Legal Documents",
    author = "T.y.s.s, Santosh  and
      Isaia, Apolline  and
      Hong, Shiyu  and
      Grabmair, Matthias",
    editor = "Al-Onaizan, Yaser  and
      Bansal, Mohit  and
      Chen, Yun-Nung",
    booktitle = "Findings of the Association for Computational Linguistics: EMNLP 2024",
    month = nov,
    year = "2024",
    address = "Miami, Florida, USA",
    publisher = "Association for Computational Linguistics",
    url = "https://aclanthology.org/2024.findings-emnlp.433/",
    doi = "10.18653/v1/2024.findings-emnlp.433",
    pages = "7357--7364"
}

@inproceedings{belfathi-selective-2025,
  TITLE = {{Is Selective Masking A Key to Improving Domain Adaptation for Masked Language Model?}},
  AUTHOR = {Belfathi, Anas and Gallina, Ygor and Hernandez, Nicolas and Monceaux, Laura and Dufour, Richard},
  URL = {https://hal.science/hal-05071803},
  BOOKTITLE = {{International Conference on Artificial Intelligence and Law}},
  ADDRESS = {Chicago, United States},
  YEAR = {2025},
  MONTH = Jun,
  HAL_ID = {hal-05071803},
  HAL_VERSION = {v1},
}

@inproceedings{dernoncourt-etal-2017-neural,
    title = "Neural Networks for Joint Sentence Classification in Medical Paper Abstracts",
    author = "Dernoncourt, Franck  and
      Lee, Ji Young  and
      Szolovits, Peter",
    editor = "Lapata, Mirella  and
      Blunsom, Phil  and
      Koller, Alexander",
    booktitle = "Proceedings of the 15th Conference of the {E}uropean Chapter of the Association for Computational Linguistics: Volume 2, Short Papers",
    month = apr,
    year = "2017",
    address = "Valencia, Spain",
    publisher = "Association for Computational Linguistics",
    url = "https://aclanthology.org/E17-2110/",
    pages = "694--700"
}

@article{gonçalves_2020,
author = {Gon\c{c}alves, S\'{e}rgio and Cortez, Paulo and Moro, S\'{e}rgio},
title = {A deep learning classifier for sentence classification in biomedical and computer science abstracts},
year = {2020},
issue_date = {Jun 2020},
publisher = {Springer-Verlag},
address = {Berlin, Heidelberg},
volume = {32},
number = {11},
issn = {0941-0643},
url = {https://doi.org/10.1007/s00521-019-04334-2},
doi = {10.1007/s00521-019-04334-2},
journal = {Neural Comput. Appl.},
month = jun,
pages = {6793–6807},
numpages = {15}
}

@article{bhattacharya2023deeprhole,
  title={DeepRhole: deep learning for rhetorical role labeling of sentences in legal case documents},
  author={Bhattacharya, Paheli and Paul, Shounak and Ghosh, Kripabandhu and Ghosh, Saptarshi and Wyner, Adam},
  journal={Artificial Intelligence and Law},
  pages={1--38},
  year={2023},
  publisher={Springer}
}

@article{curry2008looking,
  title={Looking for Law in All the Wrong Places? Foreign Law and Support for the US Supreme Court},
  author={Curry, Brett and Miller, Banks},
  journal={Politics \& Policy},
  volume={36},
  number={6},
  pages={1094--1124},
  year={2008},
  publisher={Wiley Online Library}
}

@inproceedings{song-etal-2022-supervised,
    title = "Supervised Prototypical Contrastive Learning for Emotion Recognition in Conversation",
    author = "Song, Xiaohui  and
      Huang, Longtao  and
      Xue, Hui  and
      Hu, Songlin",
    editor = "Goldberg, Yoav  and
      Kozareva, Zornitsa  and
      Zhang, Yue",
    booktitle = "Proceedings of the 2022 Conference on Empirical Methods in Natural Language Processing",
    month = dec,
    year = "2022",
    address = "Abu Dhabi, United Arab Emirates",
    publisher = "Association for Computational Linguistics",
    url = "https://aclanthology.org/2022.emnlp-main.347/",
    doi = "10.18653/v1/2022.emnlp-main.347",
    pages = "5197--5206"
}

@inproceedings{chen-etal-2023-consistent,
    title = "Consistent Prototype Learning for Few-Shot Continual Relation Extraction",
    author = "Chen, Xiudi  and
      Wu, Hui  and
      Shi, Xiaodong",
    editor = "Rogers, Anna  and
      Boyd-Graber, Jordan  and
      Okazaki, Naoaki",
    booktitle = "Proceedings of the 61st Annual Meeting of the Association for Computational Linguistics (Volume 1: Long Papers)",
    month = jul,
    year = "2023",
    address = "Toronto, Canada",
    publisher = "Association for Computational Linguistics",
    url = "https://aclanthology.org/2023.acl-long.409/",
    doi = "10.18653/v1/2023.acl-long.409",
    pages = "7409--7422"
}

@inproceedings{wu-etal-2023-mproto,
    title = "{MP}roto: Multi-Prototype Network with Denoised Optimal Transport for Distantly Supervised Named Entity Recognition",
    author = "Wu, Shuhui  and
      Shen, Yongliang  and
      Tan, Zeqi  and
      Ren, Wenqi  and
      Guo, Jietian  and
      Pu, Shiliang  and
      Lu, Weiming",
    editor = "Bouamor, Houda  and
      Pino, Juan  and
      Bali, Kalika",
    booktitle = "Proceedings of the 2023 Conference on Empirical Methods in Natural Language Processing",
    month = dec,
    year = "2023",
    address = "Singapore",
    publisher = "Association for Computational Linguistics",
    url = "https://aclanthology.org/2023.emnlp-main.145/",
    doi = "10.18653/v1/2023.emnlp-main.145",
    pages = "2361--2374"
}

@article{hochreiter1997long,
  title={Long short-term memory},
  author={Hochreiter, Sepp and Schmidhuber, J{\"u}rgen},
  journal={Neural computation},
  volume={9},
  number={8},
  pages={1735--1780},
  year={1997},
  publisher={MIT press}
}

@inproceedings{yang2016hierarchical,
  title={Hierarchical attention networks for document classification},
  author={Yang, Zichao and Yang, Diyi and Dyer, Chris and He, Xiaodong and Smola, Alex and Hovy, Eduard},
  booktitle={Proceedings of the 2016 conference of the North American chapter of the association for computational linguistics: human language technologies},
  pages={1480--1489},
  year={2016}
}

@inproceedings{mind_2019,
author = {Ming, Yao and Xu, Panpan and Qu, Huamin and Ren, Liu},
title = {Interpretable and Steerable Sequence Learning via Prototypes},
year = {2019},
isbn = {9781450362016},
publisher = {Association for Computing Machinery},
address = {New York, NY, USA},
url = {https://doi.org/10.1145/3292500.3330908},
doi = {10.1145/3292500.3330908},
booktitle = {Proceedings of the 25th ACM SIGKDD International Conference on Knowledge Discovery \& Data Mining},
pages = {903–913},
numpages = {11},
location = {Anchorage, AK, USA},
series = {KDD '19}
}

@article{Zhang_Liu_Wang_Lu_Lee_2022, title={ProtGNN: Towards Self-Explaining Graph Neural Networks}, volume={36}, url={https://ojs.aaai.org/index.php/AAAI/article/view/20898}, DOI={10.1609/aaai.v36i8.20898}, abstractNote={Despite the recent progress in Graph Neural Networks (GNNs), it remains challenging to explain the predictions made by GNNs. Existing explanation methods mainly focus on post-hoc explanations where another explanatory model is employed to provide explanations for a trained GNN. The fact that post-hoc methods fail to reveal the original reasoning process of GNNs raises the need of building GNNs with built-in interpretability. In this work, we propose Prototype Graph Neural Network (ProtGNN), which combines prototype learning with GNNs and provides a new perspective on the explanations of GNNs. In ProtGNN, the explanations are naturally derived from the case-based reasoning process and are actually used during classification. The prediction of ProtGNN is obtained by comparing the inputs to a few learned prototypes in the latent space. Furthermore, for better interpretability and higher efficiency, a novel conditional subgraph sampling module is incorporated to indicate which part of the input graph is most similar to each prototype in ProtGNN+. Finally, we evaluate our method on a wide range of datasets and perform concrete case studies. Extensive results show that ProtGNN and ProtGNN+ can provide inherent interpretability while achieving accuracy on par with the non-interpretable counterparts.}, number={8}, journal={Proceedings of the AAAI Conference on Artificial Intelligence}, author={Zhang, Zaixi and Liu, Qi and Wang, Hao and Lu, Chengqiang and Lee, Cheekong}, year={2022}, month={Jun.}, pages={9127-9135} }

@inproceedings{lai-etal-2021-learning,
    title = "Learning Prototype Representations Across Few-Shot Tasks for Event Detection",
    author = "Lai, Viet  and
      Dernoncourt, Franck  and
      Nguyen, Thien Huu",
    editor = "Moens, Marie-Francine  and
      Huang, Xuanjing  and
      Specia, Lucia  and
      Yih, Scott Wen-tau",
    booktitle = "Proceedings of the 2021 Conference on Empirical Methods in Natural Language Processing",
    month = nov,
    year = "2021",
    address = "Online and Punta Cana, Dominican Republic",
    publisher = "Association for Computational Linguistics",
    url = "https://aclanthology.org/2021.emnlp-main.427/",
    doi = "10.18653/v1/2021.emnlp-main.427",
    pages = "5270--5277"
}

@article{ahmed2020k,
  title={The k-means algorithm: A comprehensive survey and performance evaluation},
  author={Ahmed, Mohiuddin and Seraj, Raihan and Islam, Syed Mohammed Shamsul},
  journal={Electronics},
  volume={9},
  number={8},
  pages={1295},
  year={2020},
  publisher={MDPI}
}

@inproceedings{fu-etal-2023-revisiting,
    title = "Revisiting the Knowledge Injection Frameworks",
    author = "Fu, Peng  and
      Zhang, Yiming  and
      Wang, Haobo  and
      Qiu, Weikang  and
      Zhao, Junbo",
    editor = "Bouamor, Houda  and
      Pino, Juan  and
      Bali, Kalika",
    booktitle = "Proceedings of the 2023 Conference on Empirical Methods in Natural Language Processing",
    month = dec,
    year = "2023",
    address = "Singapore",
    publisher = "Association for Computational Linguistics",
    url = "https://aclanthology.org/2023.emnlp-main.677/",
    doi = "10.18653/v1/2023.emnlp-main.677",
    pages = "10983--10997"
}

@article{Bonnard_2025, title={“Steps” towards a corpus of SCOTUS opinions annotated using a Swalesian approach}, url={https://revistaiberica.org/index.php/iberica/article/view/1168}, DOI={10.17398/2340-2785.50.45}, abstractNote={&amp;lt;p&amp;gt;&amp;lt;span style=&amp;quot;font-size: 11pt; font-family: ’Times New Roman’, serif;&amp;quot;&amp;gt;In the tradition of Moreno and Swales (2018), this paper presents the creation of a manually annotated resource for supporting teaching English for Legal Purposes (ELP) and for Natural Language Processing (NLP) purposes. After justifying the use of Supreme Court of the United States opinions, we define our coding scheme by adapting the move model of rhetorical structure in specialized discourse. We describe the methodology and the implementation of the annotation campaign. We analyze how our methodology and the resulting annotation scheme diverge from those described in the literature as well as the advantages that these divergences afford. In addition to the research article, we release several supplementary materials which aim to make the process transparent and serve other researchers aiming to annotate specialized discourse with the help of machine learning techniques.&amp;amp;nbsp;&amp;lt;/span&amp;gt;&amp;lt;/p&amp;gt;}, number={50}, journal={Ibérica}, author={Bonnard, Warren and Lavissière, Mary Catherine and Belfathi, Anas and Hernandez, Nicolas and Jacquin, Christine and Monceaux-Cachard, Laura}, year={2025}, month={Dec.}, pages={45–80} }

@book{swales1990genre,
  title={Genre analysis},
  author={Swales, John M},
  year={1990},
  publisher={Cambridge university press}
}

@book{fleiss2013statistical,
  title={Statistical methods for rates and proportions},
  author={Fleiss, Joseph L and Levin, Bruce and Paik, Myunghee Cho},
  year={2013},
  publisher={john wiley \& sons}
}

@inproceedings{belfathi2023harnessing,
  title={Harnessing GPT-3.5-Turbo for Rhetorical Role Prediction in Legal Cases.},
  author={Belfathi, Anas and Hernandez, Nicolas and Monceaux, Laura},
  booktitle={JURIX},
  pages={187--196},
  year={2023}
}

@inproceedings{zheng_llm_as_judge,
 author = {Zheng, Lianmin and Chiang, Wei-Lin and Sheng, Ying and Zhuang, Siyuan and Wu, Zhanghao and Zhuang, Yonghao and Lin, Zi and Li, Zhuohan and Li, Dacheng and Xing, Eric and Zhang, Hao and Gonzalez, Joseph E and Stoica, Ion},
 booktitle = {Advances in Neural Information Processing Systems},
 editor = {A. Oh and T. Naumann and A. Globerson and K. Saenko and M. Hardt and S. Levine},
 pages = {46595--46623},
 publisher = {Curran Associates, Inc.},
 title = {Judging LLM-as-a-Judge with MT-Bench and Chatbot Arena},
 url = {https://proceedings.neurips.cc/paper_files/paper/2023/file/91f18a1287b398d378ef22505bf41832-Paper-Datasets_and_Benchmarks.pdf},
 volume = {36},
 year = {2023}
}

@article{guo2025deepseek,
  title={Deepseek-r1: Incentivizing reasoning capability in llms via reinforcement learning},
  author={Guo, Daya and Yang, Dejian and Zhang, Haowei and Song, Junxiao and Zhang, Ruoyu and Xu, Runxin and Zhu, Qihao and Ma, Shirong and Wang, Peiyi and Bi, Xiao and others},
  journal={arXiv preprint arXiv:2501.12948},
  year={2025}
}

@misc{jiang2023mistral7b,
      title={Mistral 7B}, 
      author={Albert Q. Jiang and Alexandre Sablayrolles and Arthur Mensch and Chris Bamford and Devendra Singh Chaplot and Diego de las Casas and Florian Bressand and Gianna Lengyel and Guillaume Lample and Lucile Saulnier and Lélio Renard Lavaud and Marie-Anne Lachaux and Pierre Stock and Teven Le Scao and Thibaut Lavril and Thomas Wang and Timothée Lacroix and William El Sayed},
      year={2023},
      eprint={2310.06825},
      archivePrefix={arXiv},
      primaryClass={cs.CL},
      url={https://arxiv.org/abs/2310.06825}, 
}

@article{dubey2024llama,
  title={The llama 3 herd of models},
  author={Dubey, Abhimanyu and Jauhri, Abhinav and Pandey, Abhinav and Kadian, Abhishek and Al-Dahle, Ahmad and Letman, Aiesha and Mathur, Akhil and Schelten, Alan and Yang, Amy and Fan, Angela and others},
  journal={arXiv e-prints},
  pages={arXiv--2407},
  year={2024}
}

@article{yang2025qwen3,
  title={Qwen3 technical report},
  author={Yang, An and Li, Anfeng and Yang, Baosong and Zhang, Beichen and Hui, Binyuan and Zheng, Bo and Yu, Bowen and Gao, Chang and Huang, Chengen and Lv, Chenxu and others},
  journal={arXiv preprint arXiv:2505.09388},
  year={2025}
}

@inproceedings{dettmers2023_qlora,
 author = {Dettmers, Tim and Pagnoni, Artidoro and Holtzman, Ari and Zettlemoyer, Luke},
 booktitle = {Advances in Neural Information Processing Systems},
 editor = {A. Oh and T. Naumann and A. Globerson and K. Saenko and M. Hardt and S. Levine},
 pages = {10088--10115},
 publisher = {Curran Associates, Inc.},
 title = {QLoRA: Efficient Finetuning of Quantized LLMs},
 url = {https://proceedings.neurips.cc/paper_files/paper/2023/file/1feb87871436031bdc0f2beaa62a049b-Paper-Conference.pdf},
 volume = {36},
 year = {2023}
}

@article{wang2023improving,
  title={Improving text embeddings with large language models},
  author={Wang, Liang and Yang, Nan and Huang, Xiaolong and Yang, Linjun and Majumder, Rangan and Wei, Furu},
  journal={arXiv preprint arXiv:2401.00368},
  year={2023}
}

@inproceedings{naguib-etal-2024-shot,
    title = "Few-shot clinical entity recognition in {E}nglish, {F}rench and {S}panish: masked language models outperform generative model prompting",
    author = "Naguib, Marco  and
      Tannier, Xavier  and
      N{\'e}v{\'e}ol, Aur{\'e}lie",
    editor = "Al-Onaizan, Yaser  and
      Bansal, Mohit  and
      Chen, Yun-Nung",
    booktitle = "Findings of the Association for Computational Linguistics: EMNLP 2024",
    month = nov,
    year = "2024",
    address = "Miami, Florida, USA",
    publisher = "Association for Computational Linguistics",
    url = "https://aclanthology.org/2024.findings-emnlp.400/",
    doi = "10.18653/v1/2024.findings-emnlp.400",
    pages = "6829--6852"
}

@inproceedings{chen_2019,
 author = {Chen, Chaofan and Li, Oscar and Tao, Daniel and Barnett, Alina and Rudin, Cynthia and Su, Jonathan K},
 booktitle = {Advances in Neural Information Processing Systems},
 editor = {H. Wallach and H. Larochelle and A. Beygelzimer and F. d\textquotesingle Alch\'{e}-Buc and E. Fox and R. Garnett},
 pages = {},
 publisher = {Curran Associates, Inc.},
 title = {This Looks Like That: Deep Learning for Interpretable Image Recognition},
 url = {https://proceedings.neurips.cc/paper_files/paper/2019/file/adf7ee2dcf142b0e11888e72b43fcb75-Paper.pdf},
 volume = {32},
 year = {2019}
}

@INPROCEEDINGS{yang2018robust,
  author={Yang, Hong-Ming and Zhang, Xu-Yao and Yin, Fei and Liu, Cheng-Lin},
  booktitle={2018 IEEE/CVF Conference on Computer Vision and Pattern Recognition}, 
  title={Robust Classification with Convolutional Prototype Learning}, 
  year={2018},
  volume={},
  number={},
  pages={3474-3482},
  doi={10.1109/CVPR.2018.00366}}

@article{SOURATI2023110418,
title = {Robust and explainable identification of logical fallacies in natural language arguments},
journal = {Knowledge-Based Systems},
volume = {266},
pages = {110418},
year = {2023},
issn = {0950-7051},
doi = {https://doi.org/10.1016/j.knosys.2023.110418},
url = {https://www.sciencedirect.com/science/article/pii/S0950705123001685},
author = {Zhivar Sourati and Vishnu Priya {Prasanna Venkatesh} and Darshan Deshpande and Himanshu Rawlani and Filip Ilievski and Hông-Ân Sandlin and Alain Mermoud}
}

@inproceedings{das-etal-2022-prototex,
    title = "{P}roto{TE}x: Explaining Model Decisions with Prototype Tensors",
    author = "Das, Anubrata  and
      Gupta, Chitrank  and
      Kovatchev, Venelin  and
      Lease, Matthew  and
      Li, Junyi Jessy",
    editor = "Muresan, Smaranda  and
      Nakov, Preslav  and
      Villavicencio, Aline",
    booktitle = "Proceedings of the 60th Annual Meeting of the Association for Computational Linguistics (Volume 1: Long Papers)",
    month = may,
    year = "2022",
    address = "Dublin, Ireland",
    publisher = "Association for Computational Linguistics",
    url = "https://aclanthology.org/2022.acl-long.213/",
    doi = "10.18653/v1/2022.acl-long.213",
    pages = "2986--2997"
}

@inproceedings{chalkidis-etal-2020-legal,
    title = "{LEGAL}-{BERT}: The Muppets straight out of Law School",
    author = "Chalkidis, Ilias  and
      Fergadiotis, Manos  and
      Malakasiotis, Prodromos  and
      Aletras, Nikolaos  and
      Androutsopoulos, Ion",
    booktitle = "Findings of the Association for Computational Linguistics: EMNLP 2020",
    month = nov,
    year = "2020",
    address = "Online",
    publisher = "Association for Computational Linguistics",
    doi = "10.18653/v1/2020.findings-emnlp.261",
    pages = "2898--2904"
}

@inproceedings{beltagy-etal-2019-scibert,
    title = "{S}ci{BERT}: A Pretrained Language Model for Scientific Text",
    author = "Beltagy, Iz  and
      Lo, Kyle  and
      Cohan, Arman",
    editor = "Inui, Kentaro  and
      Jiang, Jing  and
      Ng, Vincent  and
      Wan, Xiaojun",
    booktitle = "Proceedings of the 2019 Conference on Empirical Methods in Natural Language Processing and the 9th International Joint Conference on Natural Language Processing (EMNLP-IJCNLP)",
    month = nov,
    year = "2019",
    address = "Hong Kong, China",
    publisher = "Association for Computational Linguistics",
    url = "https://aclanthology.org/D19-1371/",
    doi = "10.18653/v1/D19-1371",
    pages = "3615--3620"
}

@article{kingma2014adam,
  title={Adam: A method for stochastic optimization},
  author={Kingma, Diederik P and Ba, Jimmy},
  journal={arXiv preprint arXiv:1412.6980},
  year={2014}
}

@inproceedings{bu-etal-2023-segment,
    title = "Segment-Level and Category-Oriented Network for Knowledge-Based Referring Expression Comprehension",
    author = "Bu, Yuqi  and
      Wu, Xin  and
      Li, Liuwu  and
      Cai, Yi  and
      Liu, Qiong  and
      Huang, Qingbao",
    editor = "Rogers, Anna  and
      Boyd-Graber, Jordan  and
      Okazaki, Naoaki",
    booktitle = "Findings of the Association for Computational Linguistics: ACL 2023",
    month = jul,
    year = "2023",
    address = "Toronto, Canada",
    publisher = "Association for Computational Linguistics",
    url = "https://aclanthology.org/2023.findings-acl.557/",
    doi = "10.18653/v1/2023.findings-acl.557",
    pages = "8745--8757"
}

@inproceedings{lee-etal-2021-enhancing,
    title = "Enhancing Content Preservation in Text Style Transfer Using Reverse Attention and Conditional Layer Normalization",
    author = "Lee, Dongkyu  and
      Tian, Zhiliang  and
      Xue, Lanqing  and
      Zhang, Nevin L.",
    editor = "Zong, Chengqing  and
      Xia, Fei  and
      Li, Wenjie  and
      Navigli, Roberto",
    booktitle = "Proceedings of the 59th Annual Meeting of the Association for Computational Linguistics and the 11th International Joint Conference on Natural Language Processing (Volume 1: Long Papers)",
    month = aug,
    year = "2021",
    address = "Online",
    publisher = "Association for Computational Linguistics",
    url = "https://aclanthology.org/2021.acl-long.8/",
    doi = "10.18653/v1/2021.acl-long.8",
    pages = "93--102"
}

@inproceedings{tsur-tulpan-2023-deeper,
    title = "A Deeper (Autoregressive) Approach to Non-Convergent Discourse Parsing",
    author = "Tsur, Oren  and
      Tulpan, Yoav",
    editor = "Bouamor, Houda  and
      Pino, Juan  and
      Bali, Kalika",
    booktitle = "Proceedings of the 2023 Conference on Empirical Methods in Natural Language Processing",
    month = dec,
    year = "2023",
    address = "Singapore",
    publisher = "Association for Computational Linguistics",
    url = "https://aclanthology.org/2023.emnlp-main.796/",
    doi = "10.18653/v1/2023.emnlp-main.796",
    pages = "12883--12895"
}

@inproceedings{ahrens-etal-2023-visually,
    title = "Visually Grounded Continual Language Learning with Selective Specialization",
    author = "Ahrens, Kyra  and
      Bengtson, Lennart  and
      Hee Lee, Jae  and
      Wermter, Stefan",
    editor = "Bouamor, Houda  and
      Pino, Juan  and
      Bali, Kalika",
    booktitle = "Findings of the Association for Computational Linguistics: EMNLP 2023",
    month = dec,
    year = "2023",
    address = "Singapore",
    publisher = "Association for Computational Linguistics",
    url = "https://aclanthology.org/2023.findings-emnlp.469/",
    doi = "10.18653/v1/2023.findings-emnlp.469",
    pages = "7037--7054"
}

@inproceedings{zhang-etal-2024-coarse,
    title = "A Coarse-to-Fine Prototype Learning Approach for Multi-Label Few-Shot Intent Detection",
    author = "Zhang, Xiaotong  and
      Li, Xinyi  and
      Zhang, Feng  and
      Wei, Zhiyi  and
      Liu, Junfeng  and
      Liu, Han",
    editor = "Al-Onaizan, Yaser  and
      Bansal, Mohit  and
      Chen, Yun-Nung",
    booktitle = "Findings of the Association for Computational Linguistics: EMNLP 2024",
    month = nov,
    year = "2024",
    address = "Miami, Florida, USA",
    publisher = "Association for Computational Linguistics",
    url = "https://aclanthology.org/2024.findings-emnlp.140/",
    doi = "10.18653/v1/2024.findings-emnlp.140",
    pages = "2489--2502"
}

@inproceedings{dernoncourt-lee-2017-pubmed,
    title = "{P}ub{M}ed 200k {RCT}: a Dataset for Sequential Sentence Classification in Medical Abstracts",
    author = "Dernoncourt, Franck  and
      Lee, Ji Young",
    editor = "Kondrak, Greg  and
      Watanabe, Taro",
    booktitle = "Proceedings of the Eighth International Joint Conference on Natural Language Processing (Volume 2: Short Papers)",
    month = nov,
    year = "2017",
    address = "Taipei, Taiwan",
    publisher = "Asian Federation of Natural Language Processing",
    url = "https://aclanthology.org/I17-2052/",
    pages = "308--313"
}
